\documentclass[10pt,twocolumn,letterpaper]{article}

\usepackage{cvpr} 
\usepackage{url}
\usepackage{amssymb}
\usepackage{booktabs}
\usepackage{amsmath}
\usepackage{pifont}
\usepackage{multirow}
\usepackage{adjustbox}
\usepackage[utf8]{inputenc} 
\usepackage{tabularx,booktabs,caption,array}
\usepackage{marvosym}   
\newcommand{\Envelope}{\textsuperscript{\Letter}}
\definecolor{cvprblue}{rgb}{0.21,0.49,0.74}
\usepackage[pagebackref,breaklinks,colorlinks,allcolors=cvprblue]{hyperref}

\def\paperID{12017} 
\def\confName{CVPR}
\def\confYear{2026}

\title{DGGT: Feedforward 4D Reconstruction of Dynamic Driving Scenes using Unposed Images}

\author{
Xiaoxue Chen$^{1,2}$\thanks{These authors contributed equally.} \quad \
Ziyi Xiong$^{1,2}$\footnotemark[1] \quad \
Yuantao Chen$^{1}$ \quad \
Gen Li$^{1}$ \quad \
Nan Wang$^{1}$
\\ 
Hongcheng Luo$^{2}$ \quad \
Long Chen$^{2}$ \quad \
Haiyang Sun$^{2}$\thanks{Project leader.} \quad \
Bing Wang$^{2}$ \quad \
Guang Chen$^{2}$ \quad
\\ 
Hangjun Ye$^{2}$\Envelope \ \ \ \ \
Hongyang Li$^{3}$ \quad \
Ya-Qin Zhang$^{1}$ \quad \
Hao Zhao$^{1,4}$\Envelope
\\[1ex] 
$^{1}$AIR, Tsinghua University \ \ \ \ 
$^{2}$Xiaomi EV \\
$^{3}$The University of Hong Kong \ \ \ \ 
$^{4}$Beijing Academy of Artificial Intelligence
}

\begin{document}
\maketitle

\begin{figure*}[htbp]
 \centering
 \includegraphics[height=0.20\textheight]{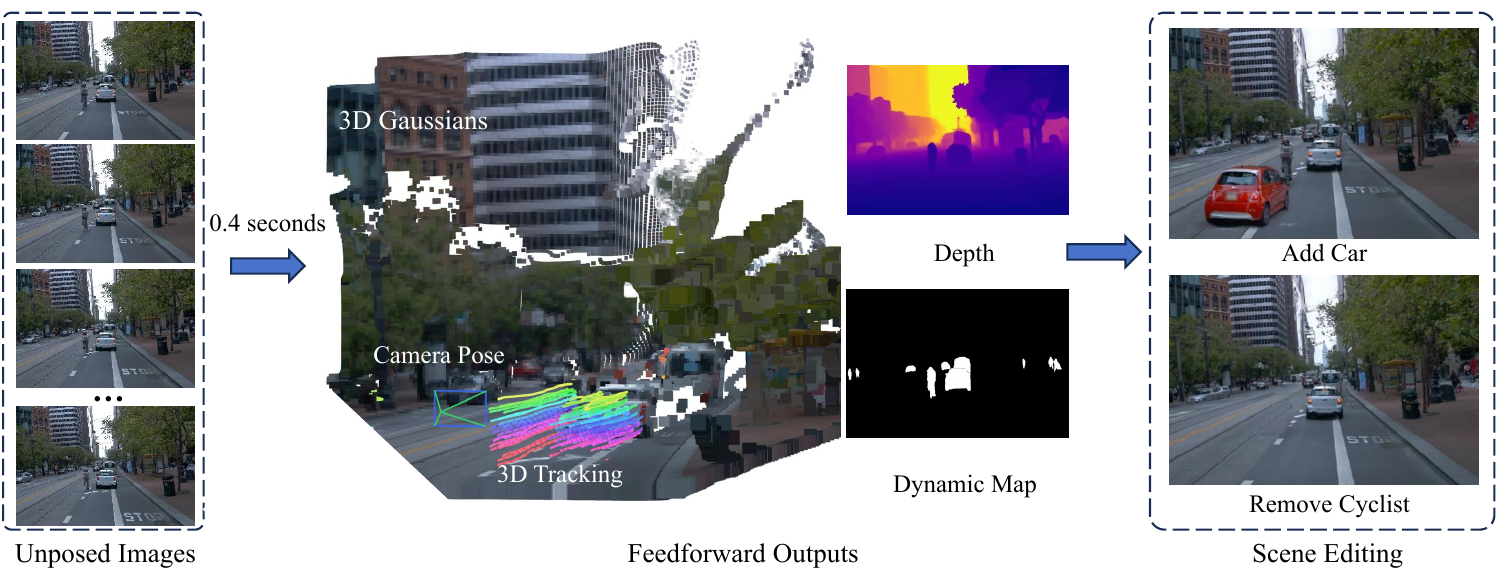}%
 \hspace{0.01\textwidth} 
  \includegraphics[width=0.28\textwidth,height=0.18\textheight,trim=5 5 5 5,clip]{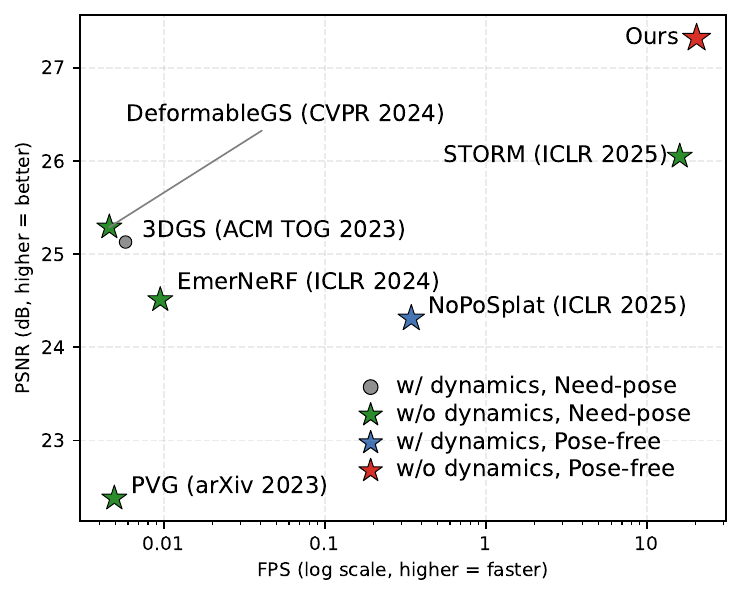}
 \caption{\textbf{Left:} Our feedforward framework reconstructs dynamic driving scenes directly from unposed images within 0.4 seconds, producing outputs such as camera pose, 3D Gaussian tracking, depth, and dynamic maps, which further enable instance-level scene editing. \textbf{Right:} Quantitative comparison shows that our method achieves state-of-the-art reconstruction quality with competitive inference speed, outperforming prior feedforward approaches in both accuracy and efficiency.(using single-view input as an example)}
 \label{fig:teaser}
\end{figure*}

\setlength{\abovedisplayskip}{3pt}   
\setlength{\belowdisplayskip}{3pt}
\setlength{\abovedisplayshortskip}{0pt}
\setlength{\belowdisplayshortskip}{0pt}

\begin{abstract}
Autonomous driving needs fast, scalable 4D reconstruction and re-simulation for training and evaluation, yet most methods for dynamic driving scenes still rely on per-scene optimization, known camera calibration, or short frame windows, making them slow and impractical. We revisit this problem from a feedforward perspective and introduce \textbf{Driving Gaussian Grounded Transformer (DGGT)}, a unified framework for pose-free dynamic scene reconstruction. We note that the existing formulations, treating camera pose as a required input, limit flexibility and scalability. Instead, we reformulate pose as an output of the model, enabling reconstruction directly from sparse, unposed images and supporting an arbitrary number of views for long sequences. Our approach jointly predicts per-frame 3D Gaussian maps and camera parameters, disentangles dynamics with a lightweight dynamic head, and preserves temporal consistency with a lifespan head that modulates visibility over time. A diffusion-based rendering refinement further reduces motion/interpolation artifacts and improves novel-view quality under sparse inputs. The result is a single-pass, pose-free algorithm that achieves state-of-the-art performance and speed. Trained and evaluated on large-scale driving benchmarks (Waymo, nuScenes, Argoverse2), our method outperforms prior work both when trained on each dataset and in zero-shot transfer across datasets, and it scales well as the number of input frames increases.
Our code and models are available at: \href{https://github.com/xiaomi-research/dggt}{\texttt{https://github.com/xiaomi-research/dggt}}
\end{abstract}
\section{Introduction}

Fast, scalable 4D reconstruction and re-simulation are important for autonomous driving. Training and evaluating driving stacks at scale requires quickly turning raw driving logs into editable and re-renderable scene representations that support what-if analysis \cite{yan2025renderworld,wu2023mars,tonderski2024neurad,li2025uniscene,jin2024tod3cap,li2025avd2,yang2023unisim,fan2025freesim,wang2024freevs,wang2023cadsim,bai2025fiffdepth,zhang2025flare,wang2025fresa,li2025mtgs,gao2025rad}. However, most dynamic-scene methods still rely on per-scene optimization \cite{shao2023tensor4d, park2021nerfies, luiten2024dynamic, wu20244d,lu2024drivingrecon}, which makes them slow and impractical. These limitations prevent the community from using reconstruction as a routine pre-processing step for learning and simulation \cite{martin2021nerf,zhang2024gaussian,kulhanek2024wildgaussians,gao2024relightable,wang2025semantic,yan2024street,zhou2024drivinggaussian,xu2025challenger,tian2023unsupervised,du2025roco}. As highlighted by our teaser (Fig.~\ref{fig:teaser}, Left), the desired capability is to reconstruct a dynamic driving scene directly from unposed images within a fraction of a second and to expose rich outputs (camera poses, 3D Gaussian tracking, depth, and dynamic maps) that further enable instance-level scene editing (e.g., \emph{add car}, \emph{remove cyclist}). At the same time, we want accuracy–efficiency tradeoff competitive with (or better than) prior art (Fig.~\ref{fig:teaser}, Right).


Feedforward reconstruction is a promising direction, but existing approaches fall short on dynamics and scalability. Recent feedforward 3D-GS pipelines (e.g., MVSplat \cite{chen2024mvsplat}, NoPoSplat \cite{ye2024no}, DepthSplat\cite{xu2025depthsplat}) deliver static reconstructions \cite{wang2025unifying,chen20254dnex}. STORM \cite{yang2024storm} takes a first step toward dynamic, feedforward driving scenes, but it still requires poses, is constrained by a fixed number of input frames (short windows), and exhibits degeneration on long sequences, making it difficult to use as a general drop-in pre-processor for large-scale logs. Moreover, existing formulations typically treat pose as an input, which limits flexibility for unposed or cross-dataset deployment. These gaps motivate a pose-free feedforward formulation that embraces dynamics and scales to arbitrary sequence lengths.


Our high-level design philosophy is to predict the entire 4D scene state in one pass while cleanly separating static background from moving entities and preserving temporal coherence. Concretely, we (i) make camera pose an output rather than a prerequisite \cite{elflein2025light3r,wu2025animateanymesh}; (ii) generate pixel-aligned Gaussian maps per frame, augmented with a lifespan parameter that modulates visibility over time to capture varying appearance; (iii) use a dynamic head to produce dense dynamic maps and a motion head to estimate 3D motion for interpolation between sparse timestamps; and (iv) plug in a single-step diffusion refinement that suppresses ghosting/disocclusion artifacts and restores fine details. This yields a single-pass, pose-free algorithm that reconstructs dynamic driving scenes from unposed images and naturally enables instance-level editing at the Gaussian level (Fig.~\ref{fig:teaser}, Left).


Our backbone is a ViT encoder that fuses DINO features with alternating attention to produce shared tokens for multiple heads: a camera head for per-frame extrinsics and intrinsics, a Gaussian head for color/position/rotation/scale/opacity, a lifespan head to control temporal visibility, a dynamic head for motion masks, a motion head for 3D displacements that align dynamic Gaussians across frames, and a lightweight sky head for distant backgrounds. Dynamic decomposition aggregates all static Gaussians across time with current-frame dynamic Gaussians, while linear pose interpolation and motion-guided mean interpolation fill intermediate timestamps. A differentiable GS renderer provides supervisory signals; the single-step diffusion module refines rendered frames using a random reference image from the input sequence, removing artifacts and closing view-coverage gaps. The design supports an arbitrary number of input views and long sequences without re-calibration or per-scene optimization.


On Waymo \cite{sun2020scalability}, our method reaches 27.41 PSNR with 0.39 s inference per scene (3 views, 20 frames per view), surpassing both optimization-based baselines (e.g., EmerNeRF, DeformableGS) and feedforward methods (e.g., NoPoSplat, DepthSplat, STORM) in accuracy while remaining competitive in speed (Fig.~\ref{fig:teaser}, Right). The approach generalizes zero-shot: trained only on Waymo, it achieves 25.31 PSNR on nuScenes \cite{caesar2020nuscenes} and 26.34 PSNR on Argoverse2 \cite{wilson2023argoverse}; training on each target dataset further boosts performance. Our 3D motion estimation attains 0.183 m EPE3D with strong Acc5/Acc10 and lower angular error than STORM, enabling consistent 3D tracking across frames. The system scales gracefully with 4/8/16-view inputs, and the lifespan and diffusion components provide clear gains. Finally, our representation enables instance-level editing (removing or shifting vehicles and inserting novel cars/cyclists from other scenes) while diffusion refinement fixes compositing holes.







\section{Related Works}

\textbf{Dynamic scene reconstruction} aims to recover a time-varying 3D representation of a scene from an image sequence. Recent extensions of NeRF \cite{li2022neural, pumarola2021d,shao2023tensor4d,park2021nerfies,mildenhall2021nerf,wang2021neus,yu2022monosdf,liu2024rip,yuan2024slimmerf} and 3DGS \cite{luiten2024dynamic, wu20244d, yan2024street,zhou2024drivinggaussian,chen2024pgsr,cheng2024gaussianpro,zhang2024drone,ye2025gs,ma2025b,miao2025evolsplat,xu2025ad,chen2025lidar,hess2025splatad} to dynamic settings have achieved high-fidelity reconstructions \cite{zhang2023copilot4d,pun2023neural,guo2025dist,song2025coda,zhao2025recondreamer++,yan2025streetcrafter,mao2025dreamdrive,lu2025infinicube,yan2024street}. Based on their motion modeling strategies, existing approaches can be broadly categorized into two main classes. The first class employs temporally conditioned representations, where time is served as an explicit input to the model. For example, PVG \cite{chen2023periodic} integrates periodic vibration-based temporal signals to predict 3DGS representations. However, such methods are often impractical for downstream tasks like object editing, as they lack disentangled, object-centric representations. The second class represents the scene as a compositional scene graph, where dynamic entities are modeled independently, typically as separate NeRFs \cite{wu2023mars} or 3DGS \cite{chen2024omnire} components. While this structure facilitates object-level manipulation, these methods often rely on external 3D annotations such as bounding boxes, which are costly to obtain. Moreover, most existing methods rely on per-scene optimization, requiring hours (Fig.~\ref{fig:teaser} right) to reconstruct each scene. To address this, we propose a fast, feedforward dynamic reconstruction method that eliminates the need for pose annotations, enabling efficient and generalizable 4D modeling across diverse scenes.

\textbf{Feedforward reconstruction} infers 3D scene representations like NeRF or 3DGS directly from input observations via a single forward pass of a trained neural network. Unlike per-scene optimization methods, feedforward approaches are designed to generalize across diverse scenes and enable real-time inference. Several methods \cite{hong2023lrm,tang2024lgm,wang2025vggt,keetha2025mapanything,jevtic2025feed} adopt vision transformers to reconstruct 3D objects from multi-view images \cite{xu20254dgt}. Subsequent works such as Flash3D \cite{szymanowicz2024flash3d}, MVSplat \cite{chen2024mvsplat}, NoPoSplat \cite{ye2024no} and DepthSplat \cite{xu2025depthsplat} extend feedforward 3D Gaussian Splatting to the scene level, though they remain limited to static environments \cite{zhang2025pansplat,chen2024omnire}.  L4GM \cite{ren2024l4gm} presents the first 4D  reconstruction model, yet the application is mainly restricted to objects. More recently, STORM \cite{yang2024storm} introduces a feedforward framework for dynamic scenes, but it is constrained by the number of input frames and struggles with long sequences. In contrast, our method supports an arbitrary number of input frames and does not require camera poses, enabling more flexible and efficient dynamic scene reconstruction.

\section{Method}

\begin{figure*}[!t]
  \centering
    \includegraphics[width=.9\linewidth]{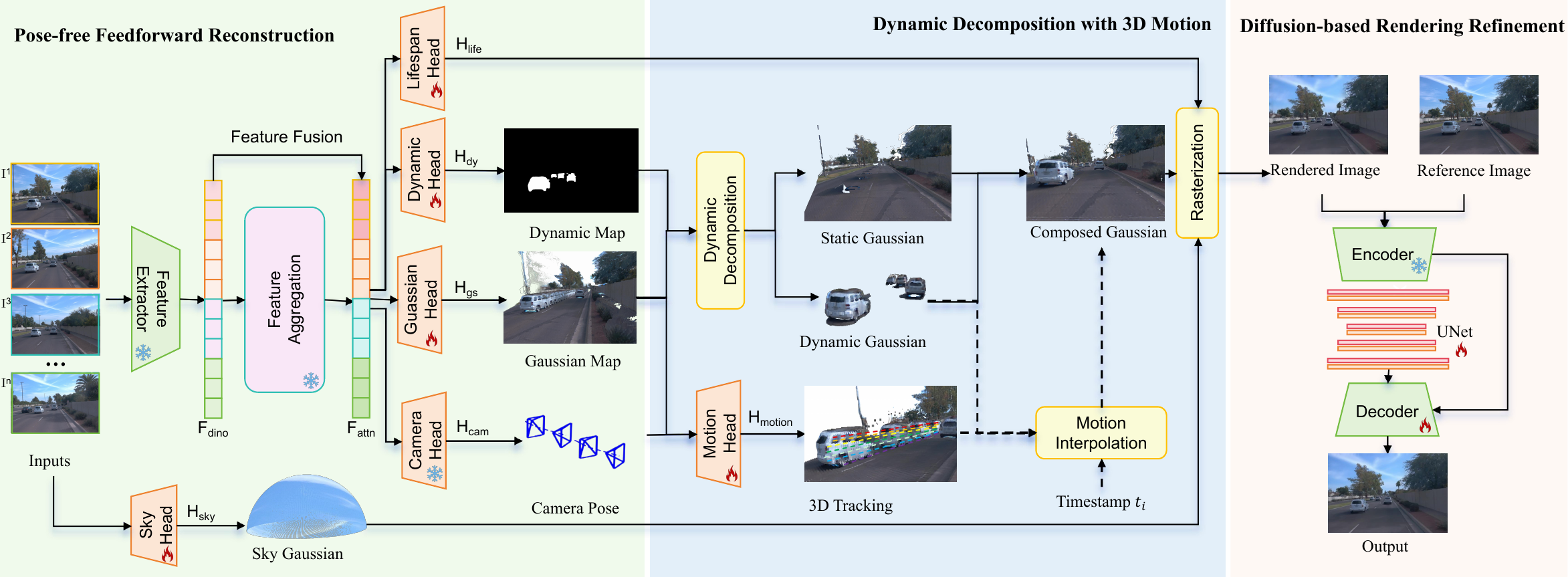}
  \caption{\textbf{Overall Architecture. } Given unposed images of dynamic scene, we estimate camera parameters, dynamic maps, and per-pixel Gaussians in a single pass. Subsequently, a motion head is employed to track dynamic objects across time, and their trajectories are interpolated to construct temporally consistent Gaussian representations. Finally, a diffusion-based rendering module refines the resulting composition, producing high-fidelity renderings.}
  \label{fig:main}
  \vspace{-.3cm}
\end{figure*}

We present a pose-free feedforward framework for reconstructing 3D scenes directly from unposed image sequences (Sec.~\ref{feedforward model}).
To address dynamic scenes under sparse observations, we introduce a motion head that predicts 3D motion and enables reliable interpolation across frames (Sec.~\ref{scene flow}).
To further improve the interpolated results, we incorporate an interpolation refinement module that enhances appearance for high-fidelity view synthesis (Sec.~\ref{fix}).
This fully feedforward design enables efficient, scalable, and high-quality 3D reconstruction without any pose inputs. The full architecture is shown in Fig.~\ref{fig:main}.

\subsection{Pose-Free Feedforward Reconstruction} \label{feedforward model}

Given a sequence of unposed RGB images $\{I^t \mid I^t \in \mathbb{R}^{H \times W \times 3}, t = 1, \ldots, N\}$ captured over $N$ timestamps of a dynamic scene, our objective is to reconstruct a temporally coherent 4D representation in a single forward pass, without requiring on external pose calibration. To this end, we propose a feedforward model $f_\theta$ that directly maps the image sequence to per-frame camera parameters $\Pi^t$ and the 3D scene representation for each timestamp. 

\paragraph{Scene Representation} We represent the scene at each frame using a pixel-aligned Gaussian map $G^t \in \mathbb{R}^{H \times W \times 15}$, where each pixel-aligned primitive at $(i, j)$ encodes RGB color $c^t_{i,j} \in \mathbb{R}^3$, 3D mean position $\mu^t_{i,j} \in \mathbb{R}^3$, rotation quaternion $r^t_{i,j} \in \mathbb{R}^4$, scale $s^t_{i,j} \in \mathbb{R}^3$,  opacity $o^t_{i,j} \in \mathbb{R}$ , and a lifespan parameter $\sigma^t_{i,j} \in \mathbb{R^{+}}$, which controls its temporal influence by modulating opacity over time. Specifically, given the predicted parameters at timestamp $t$, the opacity at another timestamp $t'$ is computed as: 
\begin{equation}
o^{t'} = o^t \cdot e^{ -\frac{1}{2} \cdot \frac{(t' - t)^2}{\sigma^t} },
\end{equation}
$\sigma$ controls the temporal spread of the Gaussian, and a larger $\sigma$ leads to a longer-lasting Gaussian in time, while a smaller $\sigma$ results in faster temporal fading. Compared with STORM \cite{yang2024storm}, this representation more accurately captures the appearance changes of static regions, resulting in a more stable dynamic decomposition.


The first frame $I^1$ is designated as the reference frame, with its camera origin serving as the world coordinate origin. All subsequent frames are aligned to this reference, ensuring consistent 3D positioning across time. 

\paragraph{Sky Modeling} While the Gaussian map captures the overall scene geometry, it is insufficient for modeling unbounded regions, especially the distant sky. To handle this, we introduce a separate set of sky Gaussians denoted by $G_{\rm sky}$. Their centers are uniformly sampled on a hemisphere of fixed radius $r_{\rm sky}$, chosen to approximate an effectively infinite background. Each sky Gaussian is assigned a fixed rotation and opacity. To determine their appearance, we project their 3D coordinates onto the input images and extract the corresponding pixel colors. These initial colors, along with the Gaussian scale parameters, are then refined by a lightweight MLP $\mathcal{H}_{\rm sky}$ to improve consistency and realism in sky modeling.

\paragraph{Model Architecture}
We adopt a ViT architecture as the backbone of our feedforward model $f_\theta$. The input images are first partitioned into patches and transformed into token sequences, which are subsequently encoded by a DINO-pretrained feature extractor~\cite{zhang2022dino} to obtain rich visual representations, denoted as $F_{\text{dino}}$. These features are then refined through an alternating-attention mechanism, yielding $F_{\text{attn}}$. The resulting features are processed by multiple prediction heads: 
the camera head $\mathcal{H}_{\text{cam}}$ for estimating camera poses, 
the Gaussian head $\mathcal{H}_{\text{gs}}$ for generating 3D Gaussians, 
and the lifespan head $\mathcal{H}_{\text{life}}$ for lifespan parameters.

During training, we leverage pretrained priors by freezing the feature extractor and camera head, while training the remaining heads from scratch. Feeding $F_{\text{attn}}$ into the prediction heads, the camera pose is obtained as $\Pi^t = \mathcal{H}_{\text{cam}}(F_{\text{attn}})$. However, we observe that $F_{\text{attn}}$ primarily encodes high-level semantics and lacks sufficient detail for appearance reconstruction. To mitigate this issue, we fuse $F_{\text{attn}}$ with the original DINO features, thereby enhancing spatial fidelity. Denoted by the arrow between $F_{\text{dino}}$ and $F_{\text{attn}}$ in Fig.~\ref{fig:main} left, the fused feature is then used by the Gaussian head to generate the Gaussian splatting map as 
    $G^t = \mathcal{H}_{\text{gs}}(F_{\text{dino}}, F_{\text{attn}}).$

This unified framework enables reconstruction of the 3D scene structure across all timestamps. Nevertheless, in dynamic environments, naïvely aggregating Gaussian maps across time leads to incoherent reconstructions due to object motion. To address this, we introduce a motion field that explicitly models dynamic object trajectories, as detailed in the following section.


\subsection{Dynamic Decomposition with 3D Motion} \label{scene flow}

Given the predicted Gaussian maps, directly aggregating them over time would produce ghosting artifacts due to moving objects, as illustrated in Fig.~\ref{fig:main} (Gaussian map). Therefore, we extend the feedforward model with a dynamic head $\mathcal{H}_{\text{dy}}$ that predicts the probability of dynamic regions, formulated as $M_d^t = \mathcal{H}_{\text{dy}}(F_{\text{attn}})$. This dynamic map distinguishes moving objects from the static background, enabling us to decompose each Gaussian map $G^t$ into a static component $G_s^t$ and a dynamic component $G_d^t$:
\begin{equation}
G_s^t = G^t \odot (1 - M_d^t), \quad G_d^t = G^t \odot M_d^t,
\end{equation}
where $\odot$ denotes element-wise multiplication. For each timestamp $t$, the full Gaussian representation is constructed by combining the sky Gaussian, the static components from all frames, and the dynamic component of the current frame as:
\begin{equation}
\hat{G}^t = \left( \bigcup_{t'=1}^N G_s^{t'} \right) \cup G_d^t \cup G_{\text{sky}}. \label{eq: aggregating} 
\end{equation}
here $G_d^t$ denotes the dynamic Gaussians at frame $t$, and $\bigcup_{t'=1}^N G_s^{t'}$ represents the union of static Gaussians from all frames. The rendered image at timestamp $t$ is then obtained via the differentiable renderer, defined as:
\begin{equation}
\hat{I}^t={\rm Renderer}(\hat{G}^t ,\Pi^t),
\end{equation}
where $\Pi^t$ denotes the predicted camera parameters.

\paragraph{3D Motion Estimation} 
In practice, the input timestamps are sparse. For an intermediate time $t_i$ with $t_i \notin \{1, \ldots, N\}$, corresponding dynamic Gaussian representation is not directly available. Therefore, it is necessary to model the motion of dynamic objects over time. Unlike prior methods such as STORM \cite{yang2024storm}, which model only Gaussian velocities, we explicitly model full 3D motion trajectories for each pixel. This richer motion representation captures complex non-linear dynamics more accurately and enables the synthesis of dynamic Gaussians at arbitrary intermediate timestamps. We exploit correspondences between input images and introduce a motion head that predicts 3D motion for a set of query pixels $\mathcal{Q} \in \mathbb{R}^{q \times 2}$, where $q$ is the number of queries. Concretely, for any pair of timestamps $t_a, t_b \in \{1, \ldots, N\}$, 
the motion head estimates 3D motion
$F(t_a, t_b) \in \mathbb{R}^{q \times 3}$, which provides per-pixel 3D displacement vectors, enabling temporal alignment of Gaussian primitives across frames.

Specifically, we introduce a transformer-based motion head $\mathcal{H}_{\mathrm{motion}}$, which jointly processes 2D images and 3D points extracted from Gaussian maps. The images are first encoded into multi-scale features, which are then associated with 3D points to construct a spatio-temporal feature cloud. For each timestamp $t_a$, we select the one-valued pixels in the dynamic map $M_d^{t_a}$ as query pixels $\mathcal{Q}$, back-project them to initialize their 3D positions, and iteratively refine their trajectories via neighborhood-to-neighborhood attention. Formally, the motion head is defined as:
\begin{equation}
F(t_a, t_b) = \mathcal{H}_{\mathrm{motion}}(\mathcal{Q} \mid G^{t_a}, G^{t_b}, I^{t_a}, I^{t_b}),
\end{equation}
where $G^{t_a}$ and $G^{t_b}$ are the Gaussian maps and $I^{t_a}$, $I^{t_b}$ are the corresponding images at times $t_a$ and $t_b$. The motion head is initialized with pretrained weights~\cite{zhang2025tapip3d} and finetuned using a photometric loss on interpolated frames.

\begin{table*}[t]
  \centering
  \small
  \begin{tabular}{@{}lcccccc@{}} 
    \toprule
    & \multicolumn{3}{c}{\textbf{Render Quality}} &
      \multirow{2}{*}{\begin{tabular}{@{}c@{}}\textbf{Inference time (s)}\end{tabular}} &
      \multirow{2}{*}{\begin{tabular}{@{}c@{}}\textbf{Dynamic}\end{tabular}} &
      \multirow{2}{*}{\begin{tabular}{@{}c@{}}\textbf{Pose-free}\end{tabular}} \\
    \cmidrule(lr){2-4}
    \textbf{Method} & \textbf{PSNR} $\uparrow$ & \textbf{SSIM} $\uparrow$ & \textbf{D-RMSE} $\downarrow$ & & & \\
    \midrule
    EmerNeRF \cite{yang2023emernerf} &  24.51 &  0.738 & 33.99 & 14min & \ding{51} & \ding{55} \\
    3DGS \cite{kerbl20233d} & 25.13 & 0.741 & 19.68 & 23min & \ding{55} & \ding{55} \\
    PVG \cite{chen2023periodic} & 22.38 & 0.661 & 13.01 & 27min & \ding{51} & \ding{55} \\
    DeformableGS \cite{yang2024deformable} & 25.29 & 0.761 & 14.79 & 29min & \ding{51} & \ding{55} \\
    LGM \cite{tang2024lgm} & 18.53 & 0.447 & 9.07 & \underline{0.06s}  & \ding{55} & \ding{55} \\
    GS-LRM \cite{zhang2024gs} & 25.18 & 0.753 & 7.94 & \textbf{0.02s}  & \ding{55} & \ding{55} \\
    
    MVSplat \cite{chen2024mvsplat}       & 20.56 & 0.697 & 10.13 & 0.08s & \ding{55} & \ding{55} \\
    NoPoSplat \cite{ye2024no}            & 24.31 & 0.751 & 9.08 & 23.22s & \ding{55} & \ding{51} \\
    DepthSplat \cite{xu2025depthsplat}   & 23.26 & 0.696 & 10.05 & 0.11s & \ding{55} & \ding{55} \\
    STORM* \cite{yang2024storm} & 26.05 & \underline{0.819} & 5.91 & 0.50s & \ding{51} & \ding{55} \\
    
    STORM \cite{yang2024storm} & \underline{26.38} & 0.794 & 5.48 & 0.18s & \ding{51} & \ding{55} \\
    VGGT++ \cite{wang2025vggt} & 22.50 & 0.749 & \underline{3.80} & 0.24s & \ding{55} & \ding{51} \\
    \midrule
    Ours    & \textbf{27.41} & \textbf{0.846} & \textbf{3.47} & 0.39s & \ding{51} & \ding{51} \\
    \bottomrule
  \end{tabular}
  \caption{\textbf{Quantitative comparison on the Waymo dataset}. Higher PSNR and SSIM, and lower D-RMSE indicate better performance. All methods were evaluated using three input views, where the task is to interpolate the intermediate frame between adjacent views. (* denotes results from our replication). Detailed experimental settings are provided in Appendix.}
  \label{tab:4d recon}
\end{table*}

\paragraph{Motion Interpolation} With the 3D motion, we can estimate the dynamic Gaussian representation at an intermediate time $t_i \in [t_a, t_b]$, where $t_a$ and $t_b$ are adjacent timestamps. Specifically, we interpolate the mean coordinates $\mu^t_d$ of the dynamic Gaussians using the  motion prediction as:
\begin{align}
\mu^{t_i}_d = \mu^{t_a}_d + \omega^{t_i} \cdot F(t_a, t_b) , \quad
\omega^{t_i} = \frac{{t_i} - t_a}{t_b - t_a},
\end{align}
where $\omega^{t_i}$ is a linear interpolation weight and $M_d^{t_a}$ denotes the dynamic mask at $t_a$. This yields the dynamic Gaussians $G_d^{t_i}$ at the intermediate timestamp $t_i$, which are then combined as Eq.~\ref{eq: aggregating} to obtain the full Gaussian representation.  For camera pose $\Pi^{t_i}$ at timestamp $t_i$, we linearly interpolated the translation between $\Pi^{t_a}$ and $\Pi^{t_b}$, while the rotation was interpolated using spherical linear interpolation (SLERP) on quaternions.

Notably, our model predicts all camera parameters, Gaussian representations, 3D motion, and dynamic masks in a single forward pass, enabling simultaneous 4D reconstruction and scene understanding. The full forward process is defined as:
\begin{equation}
G_{\text{sky}}, \{G^t, F^t, M_d^t, \Pi^t\}_{t=1}^N = f_\theta(\{I^t\}_{t=1}^N),
\end{equation}
where $G_{\text{sky}}, G^t,F^t, M_d^t, \Pi^t$ are the sky Gaussian, Gaussian map, 3D motion, dynamic map and camera pose respectively, $f_\theta$ denotes the feedforward network, and $\{I^t\}_{t=1}^N$ is the set of input images.

\paragraph{Training Objectives} We train the feedforward reconstruction model including the motion estimation head, in an end-to-end manner. For each training iteration, we randomly sample $N \in [4,8]$ input images along with $2N$ ground-truth target images. The model then predicts $2N$ interpolated frames, which are supervised by the corresponding $2N$ ground-truth images. 
The reconstruction loss combines an $\ell_2$ term with a perceptual LPIPS term:

\begin{equation}
\mathcal{L}_{\text{rgb}} = \mathcal{L}_{\ell_2} + \lambda_{\text{LPIPS}}\mathcal{L}_{\text{LPIPS}}.
\end{equation}
To supervise opacity and dynamic maps, we adopt binary cross-entropy losses as:
\begin{equation}
\text{\scriptsize$
\mathcal{L}_{\text{opacity}} = \mathrm{BCE}(M_{\text{sky}}, \hat{M}_{\text{sky}}), \quad
\mathcal{L}_{\text{dynamic}} = \mathrm{BCE}(M_{d}, \hat{M}_{\text{dynamic}})
$}
\end{equation}

where $M_{\text{sky}}$ is derived from the rendered opacity with ground-truth $\hat{M}_{\text{sky}}$, and $M_d$ denotes the predicted dynamic map with ground-truth $\hat{M}_{\text{dynamic}}$. We impose an $\ell_1$ regularization for lifespan parameters under the assumption that most of the scene is static as $\mathcal{L}_{\text{lifespan}} = \left| \tfrac{1}{\sigma} \right|_1$.

The overall training objective is a weighted combination of these components:
\begin{equation}
\!\!\text{\scriptsize$
\mathcal{L}_{\text{feedforward}} =
\mathcal{L}_{\text{rgb}}
+ \lambda_{\text{opacity}}\mathcal{L}_{\text{opacity}}
+ \lambda_{\text{dynamic}}\mathcal{L}_{\text{dynamic}}
+ \lambda_{\text{lifespan}}\mathcal{L}_{\text{lifespan}}
$}
\end{equation}

\begin{figure*}[!t]
  \centering
  \begin{picture}(0,0)
  \end{picture}

  \begin{tabular}{c@{\hskip 0.02in} c@{\hskip 0.02in} c@{\hskip 0.02in} c@{\hskip 0.02in} c@{\hskip 0.02in} c@{\hskip 0.02in}}
    \textbf{GT} & \textbf{MVSplat} & \textbf{DepthSplat} & \textbf{NoPoSplat} & \textbf{STORM} & \textbf{Ours} \\
    
    \includegraphics[width=0.16\linewidth]{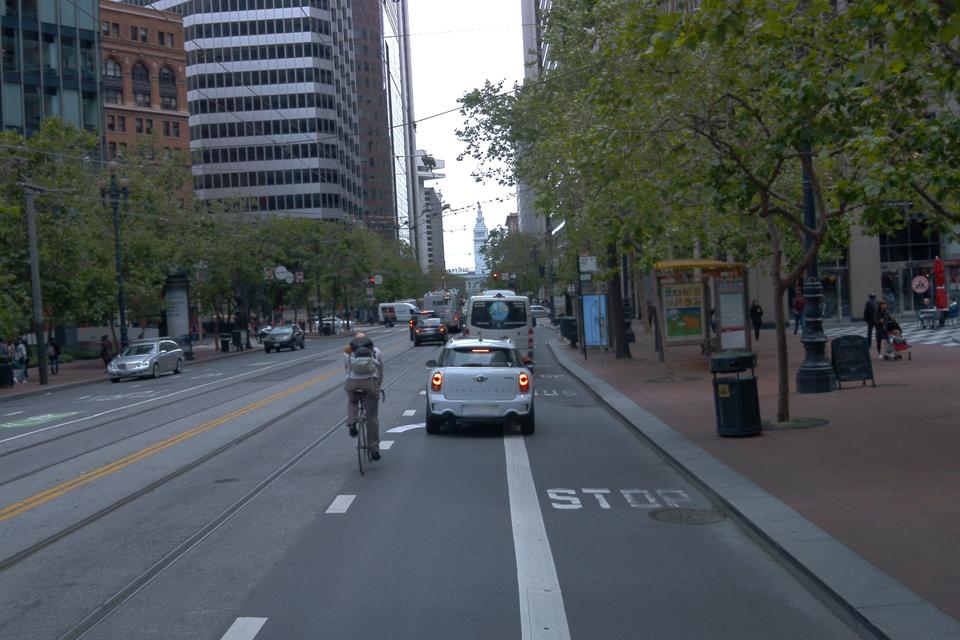} &
    \includegraphics[width=0.16\linewidth]{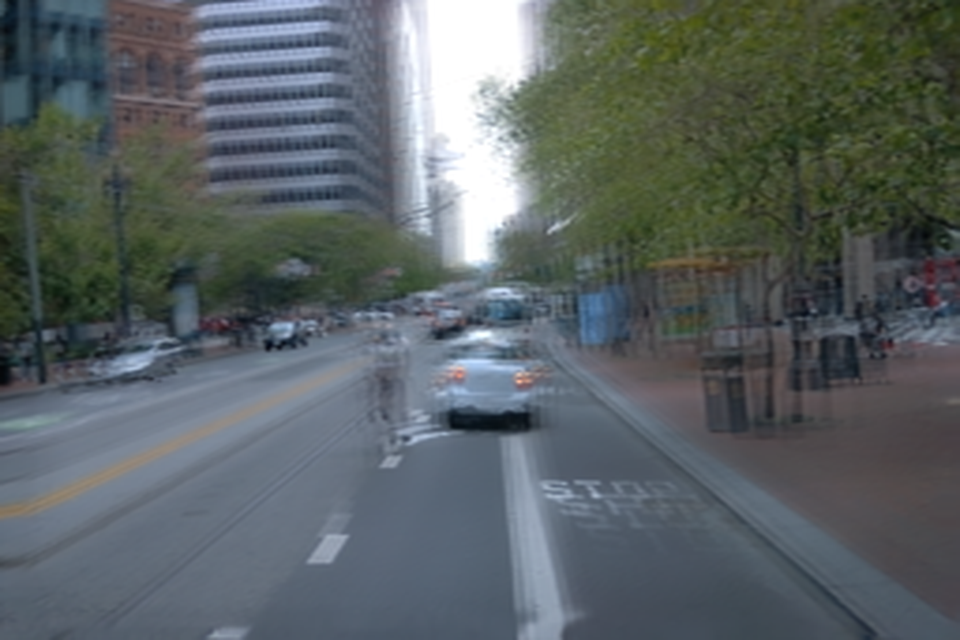} &
    \includegraphics[width=0.16\linewidth]{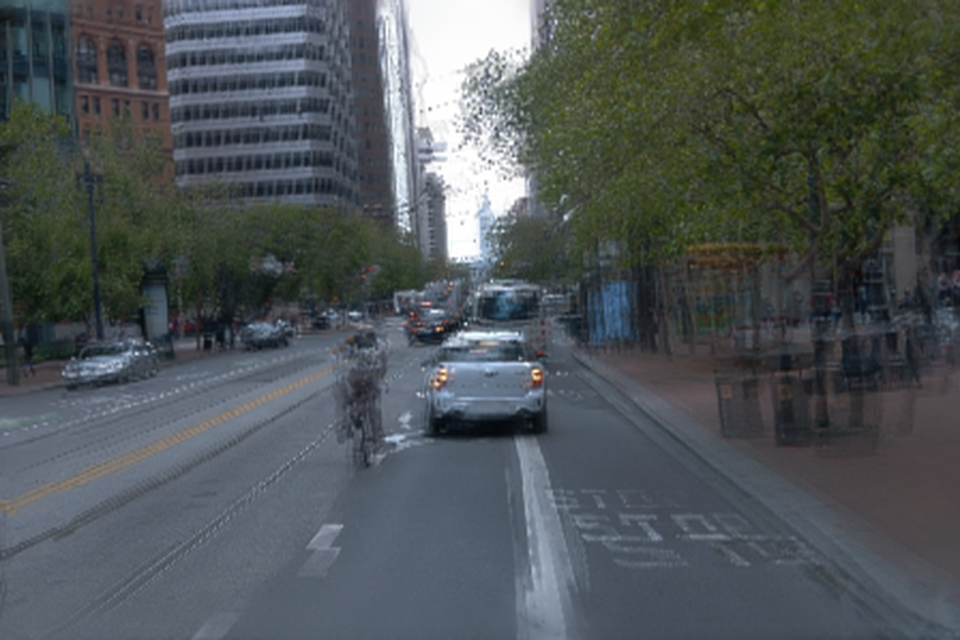} &
    \includegraphics[width=0.16\linewidth]{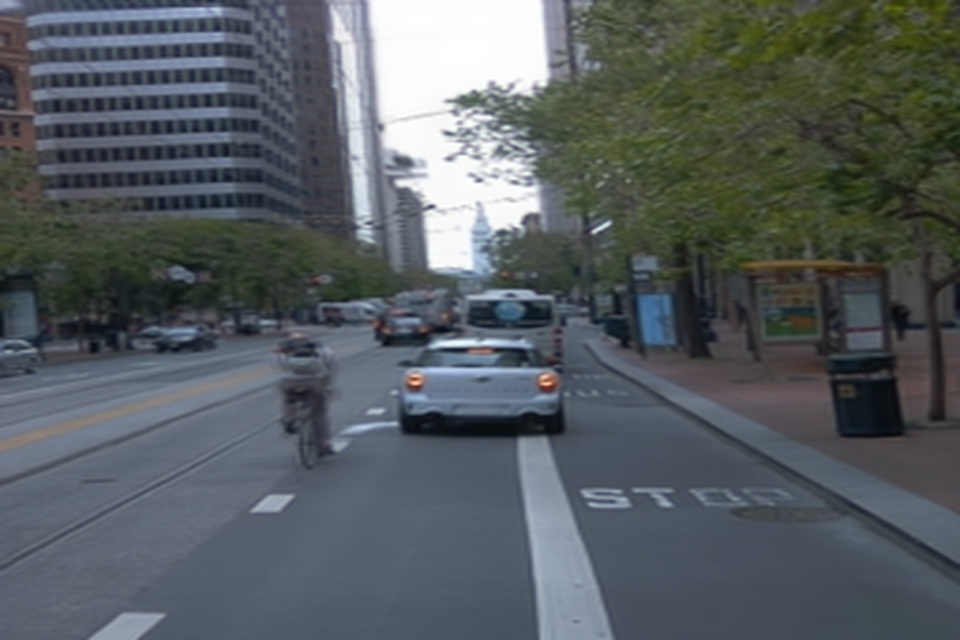} &
    \includegraphics[width=0.16\linewidth]{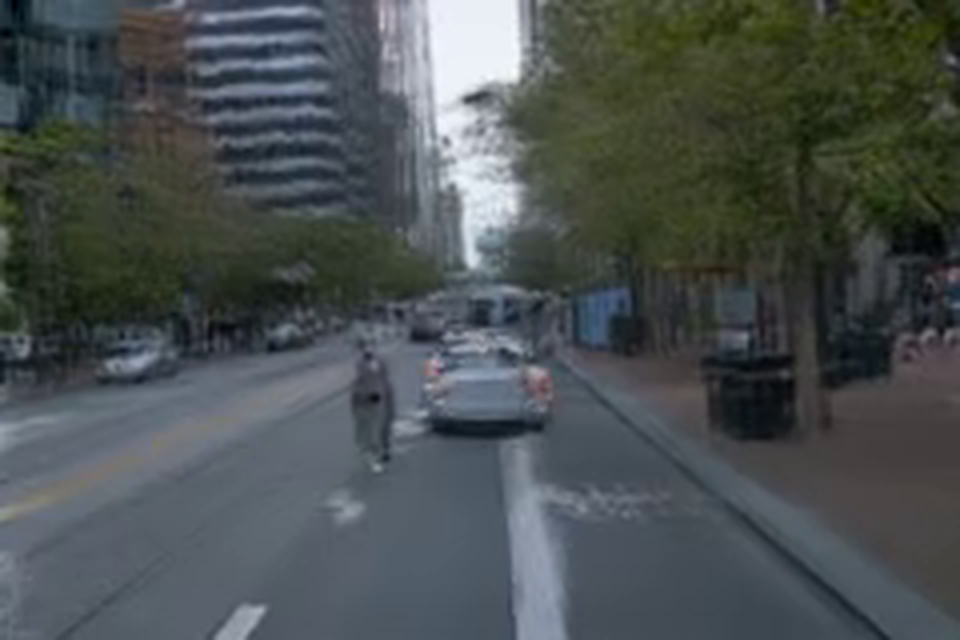} &
    \includegraphics[width=0.16\linewidth]{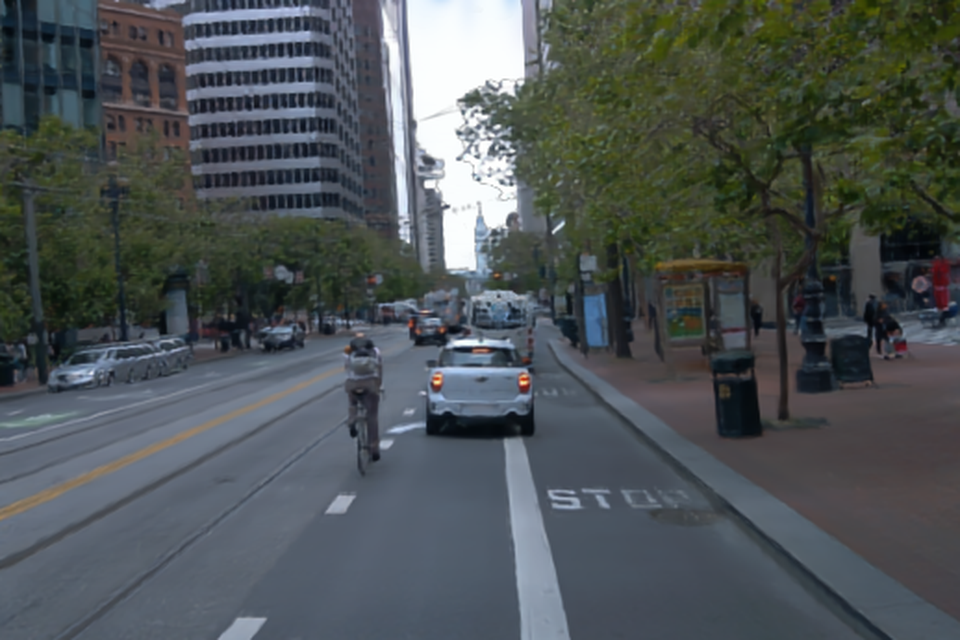} \\ [-0.5mm]
    
    \includegraphics[width=0.16\linewidth]{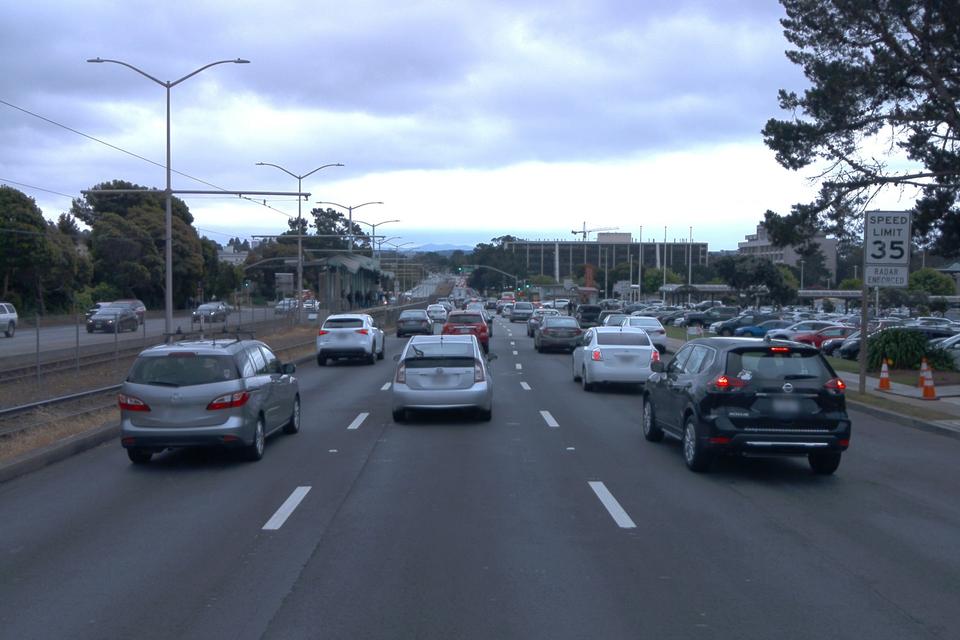} &
    \includegraphics[width=0.16\linewidth]{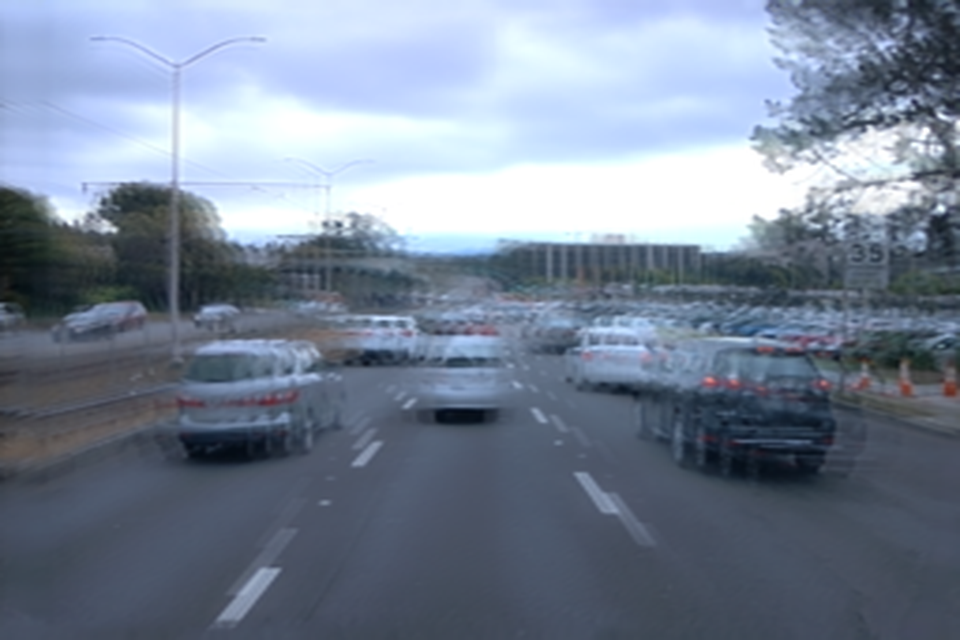} &
    \includegraphics[width=0.16\linewidth]{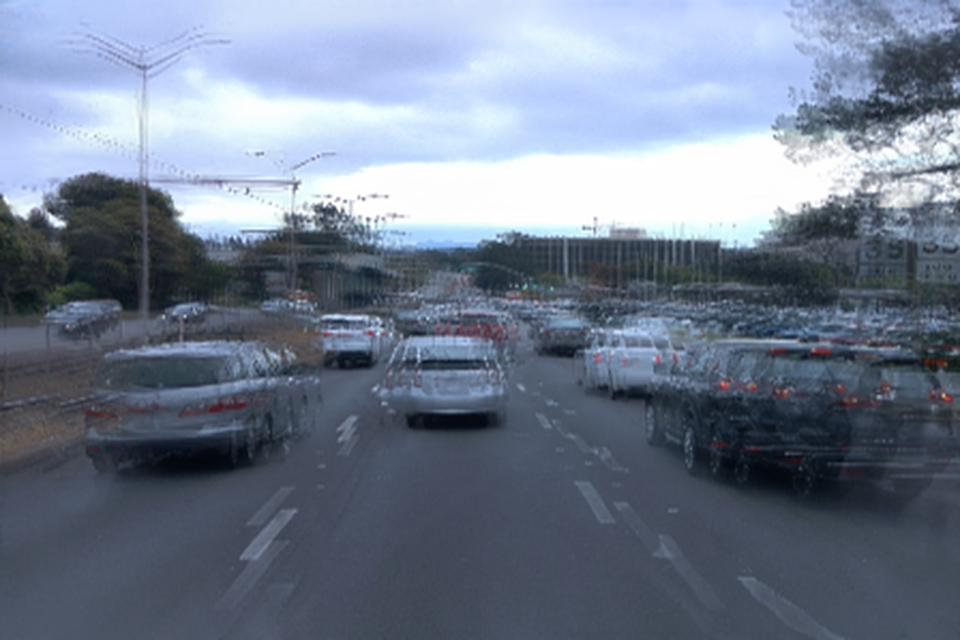} &
    \includegraphics[width=0.16\linewidth]{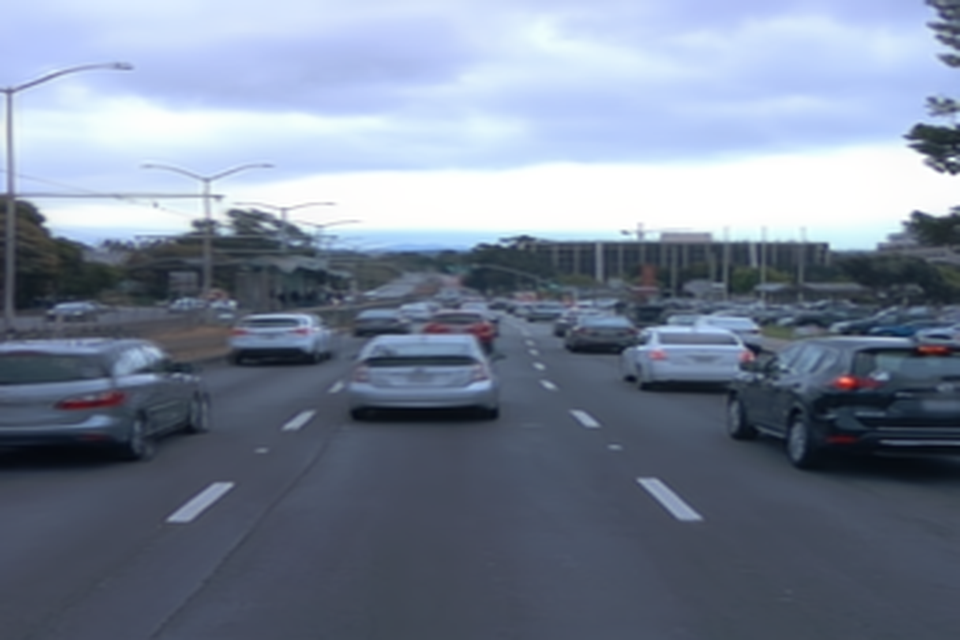} &
    \includegraphics[width=0.16\linewidth]{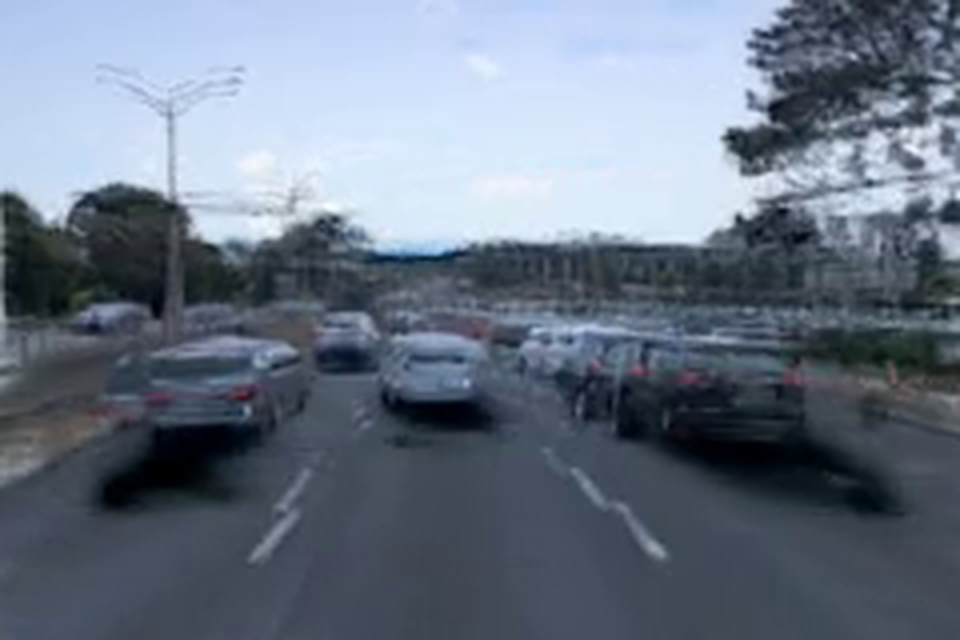} &
    \includegraphics[width=0.16\linewidth]{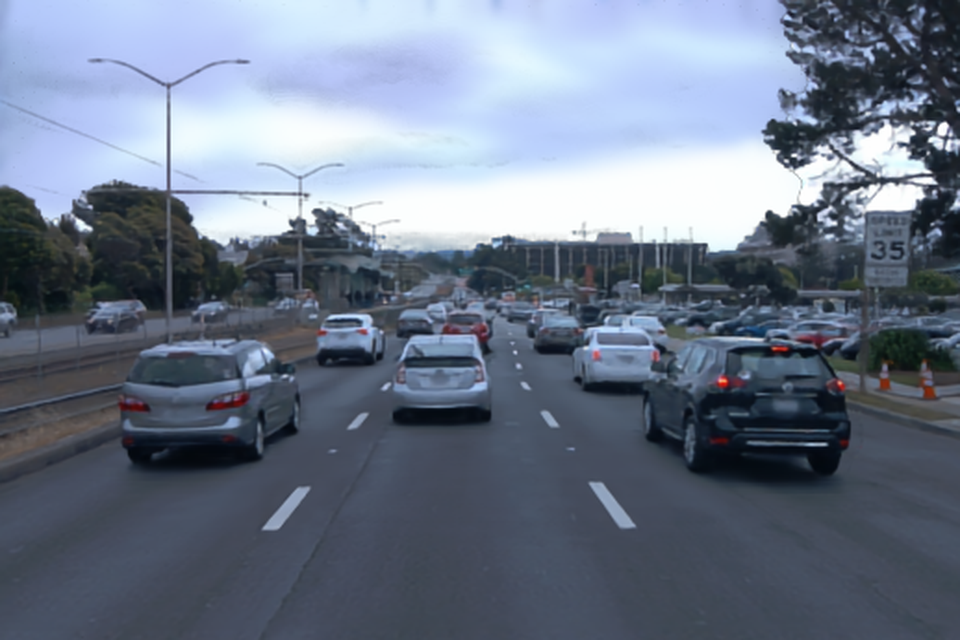} \\ [-0.5mm]
    
    \includegraphics[width=0.16\linewidth]{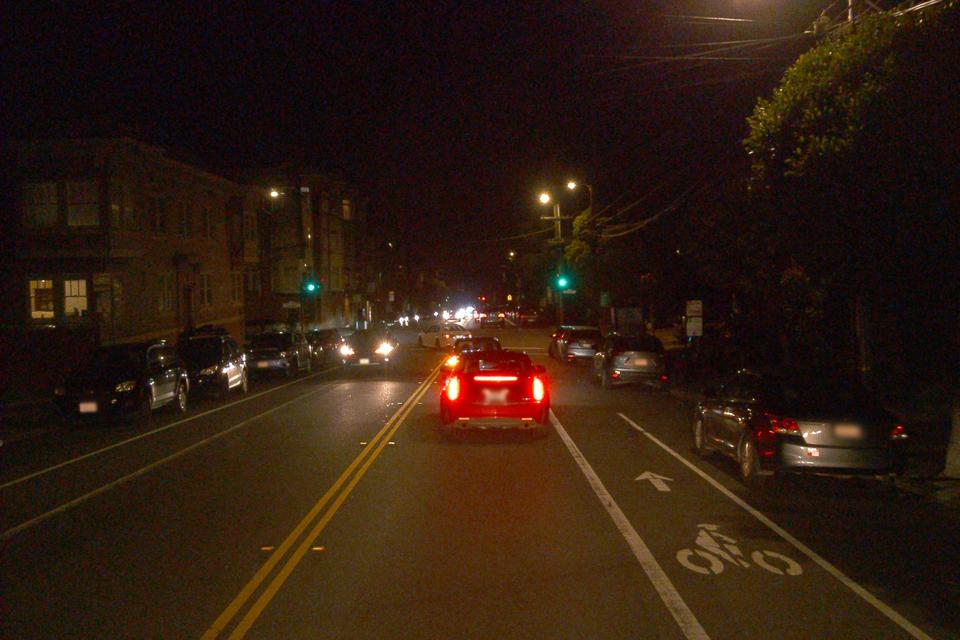} &
    \includegraphics[width=0.16\linewidth]{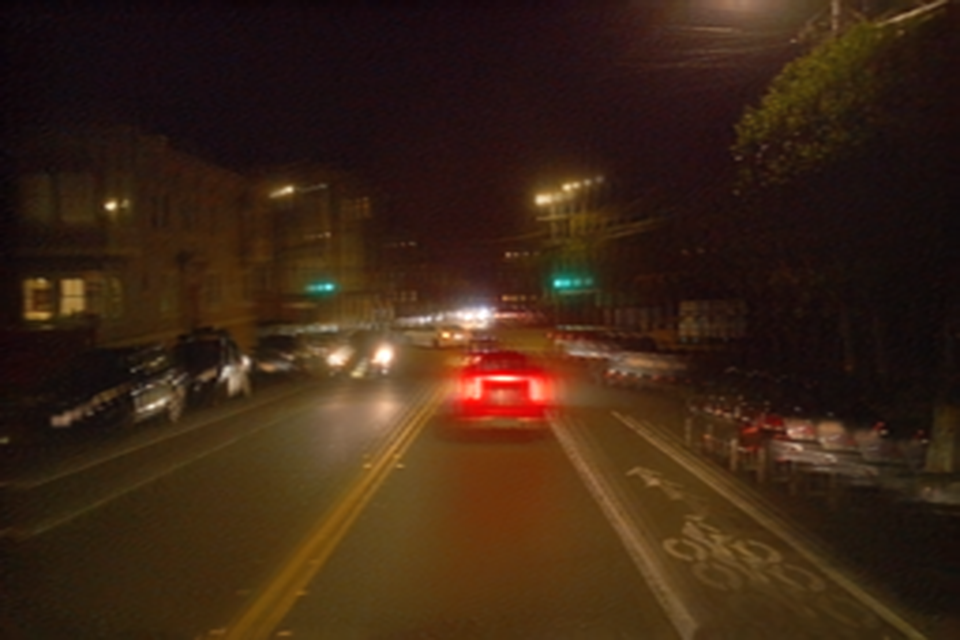} &
    \includegraphics[width=0.16\linewidth]{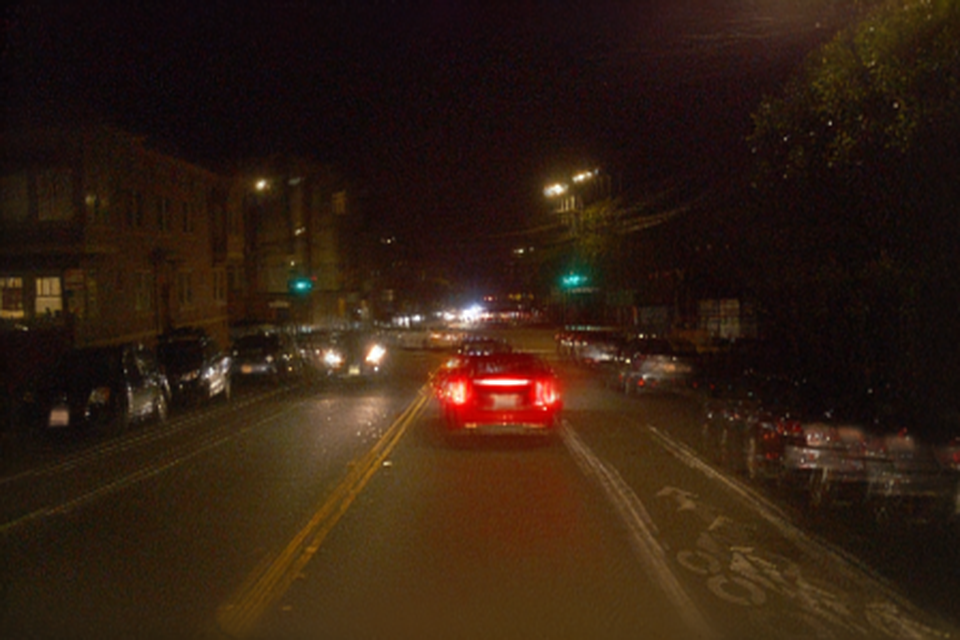} &
    \includegraphics[width=0.16\linewidth]{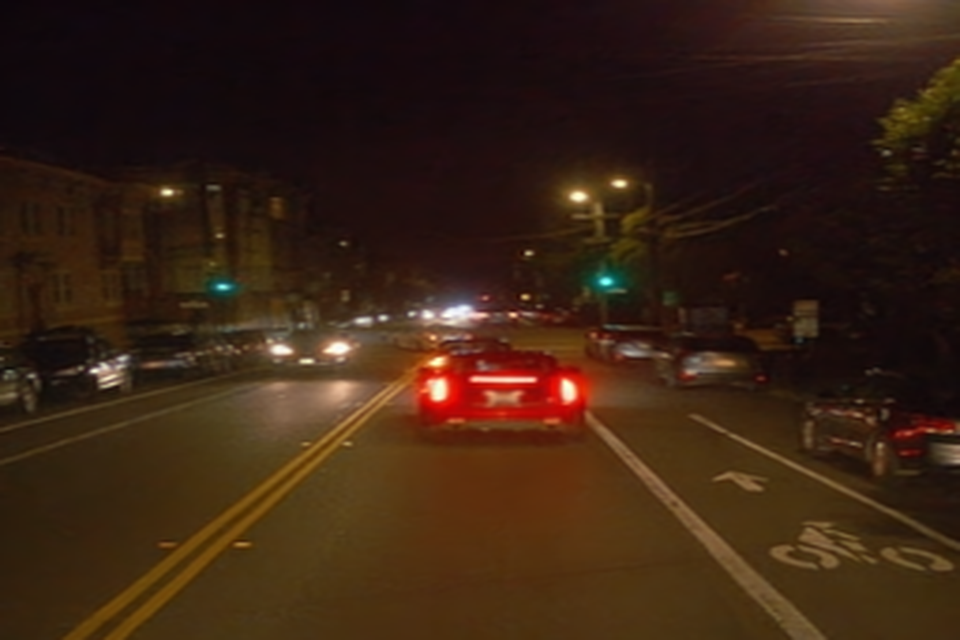} &
    \includegraphics[width=0.16\linewidth]{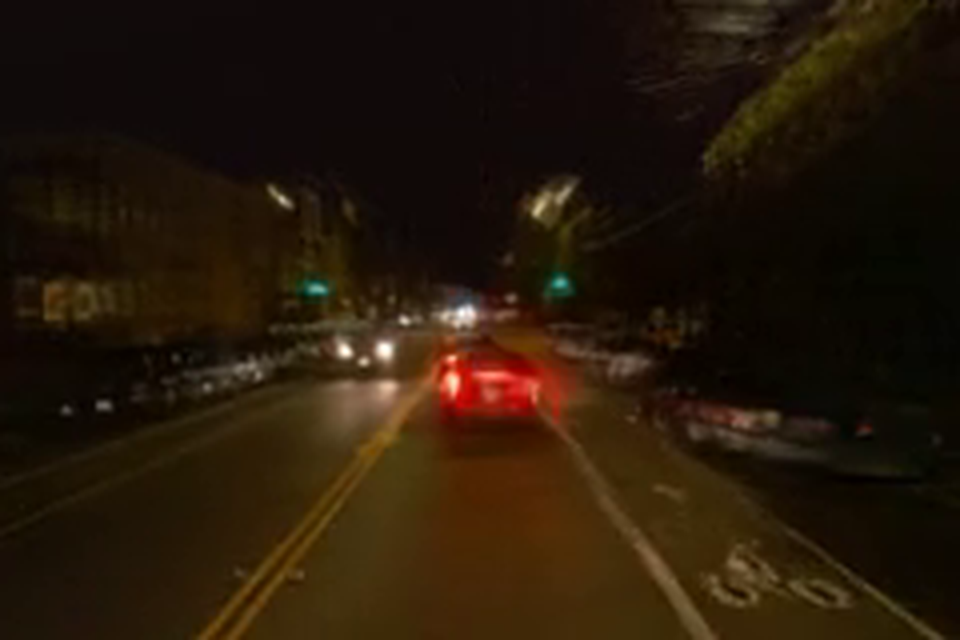} &
    \includegraphics[width=0.16\linewidth]{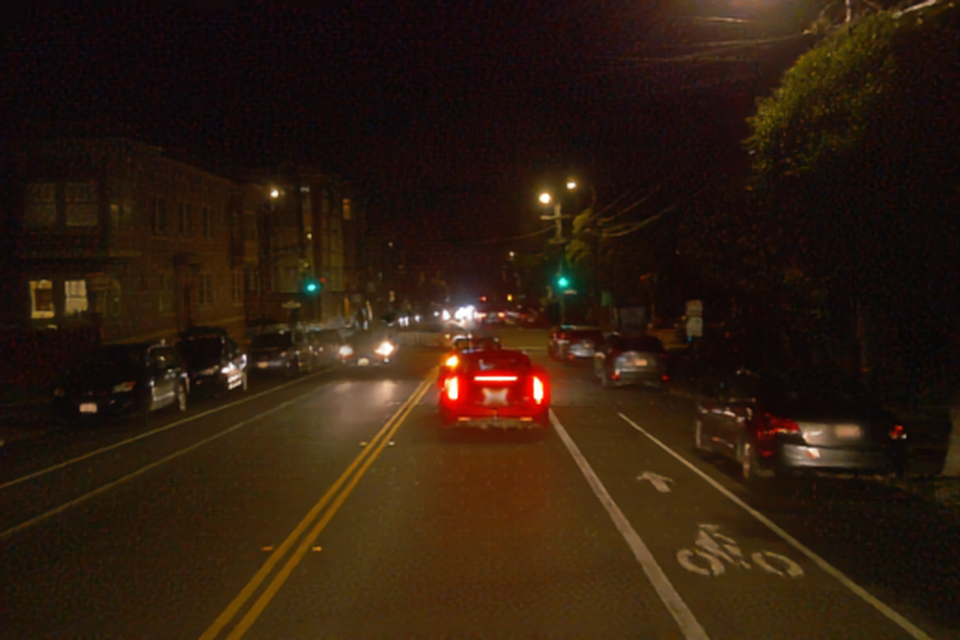} \\ [-0.5mm]
    
    \includegraphics[width=0.16\linewidth]{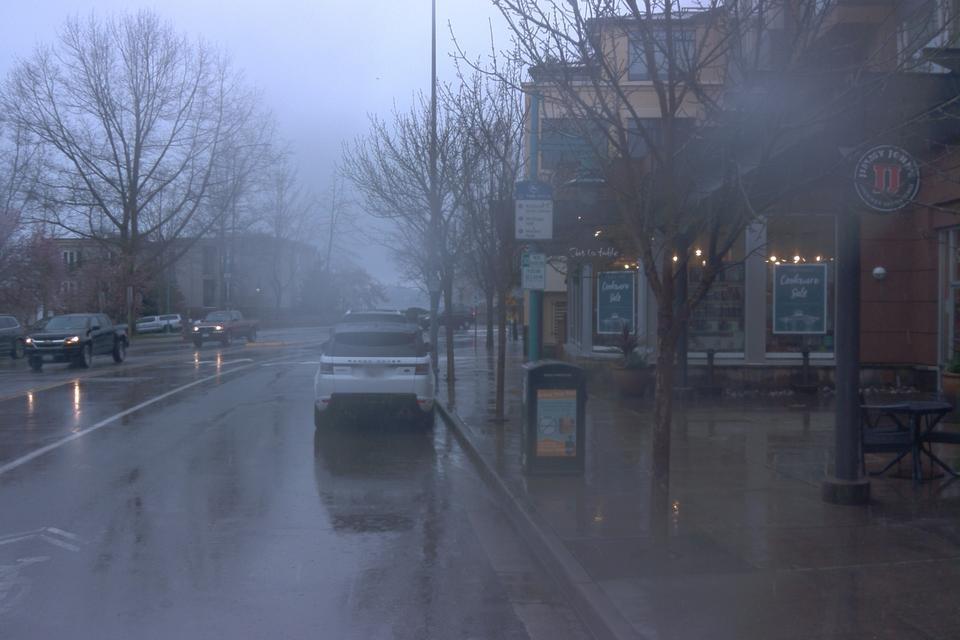} &
    \includegraphics[width=0.16\linewidth]{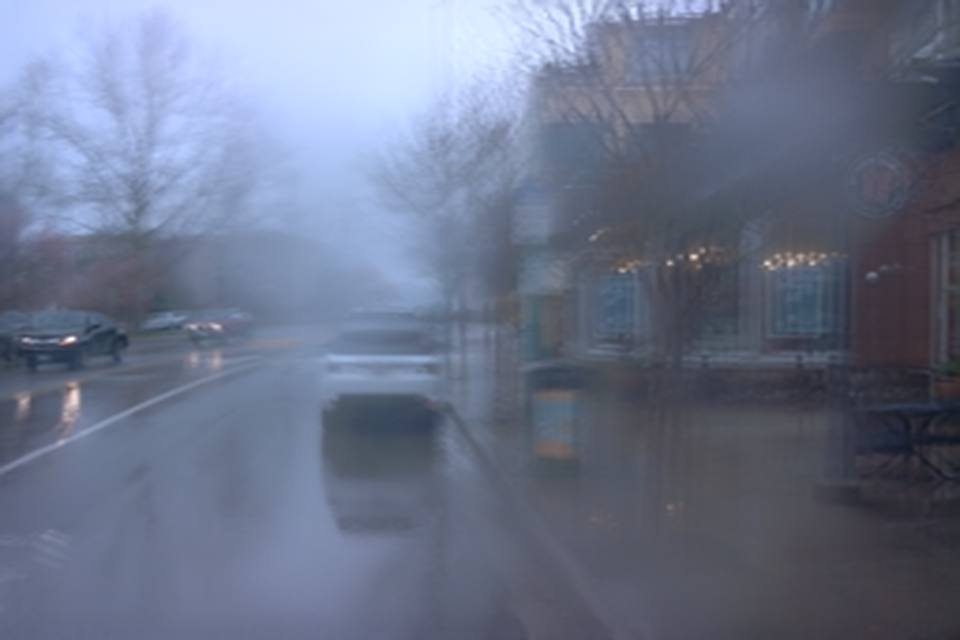} &
    \includegraphics[width=0.16\linewidth]{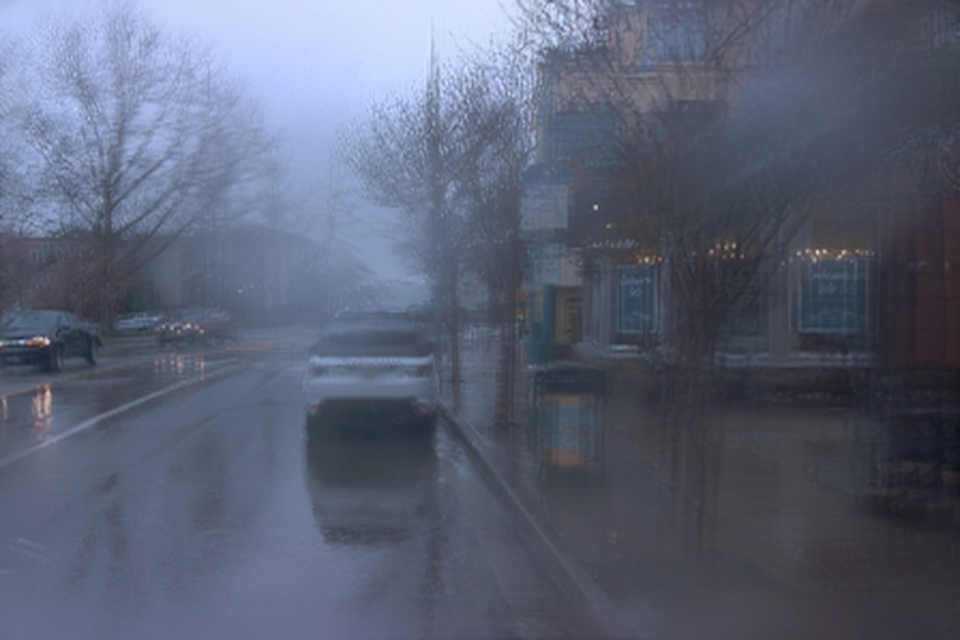} &
    \includegraphics[width=0.16\linewidth]{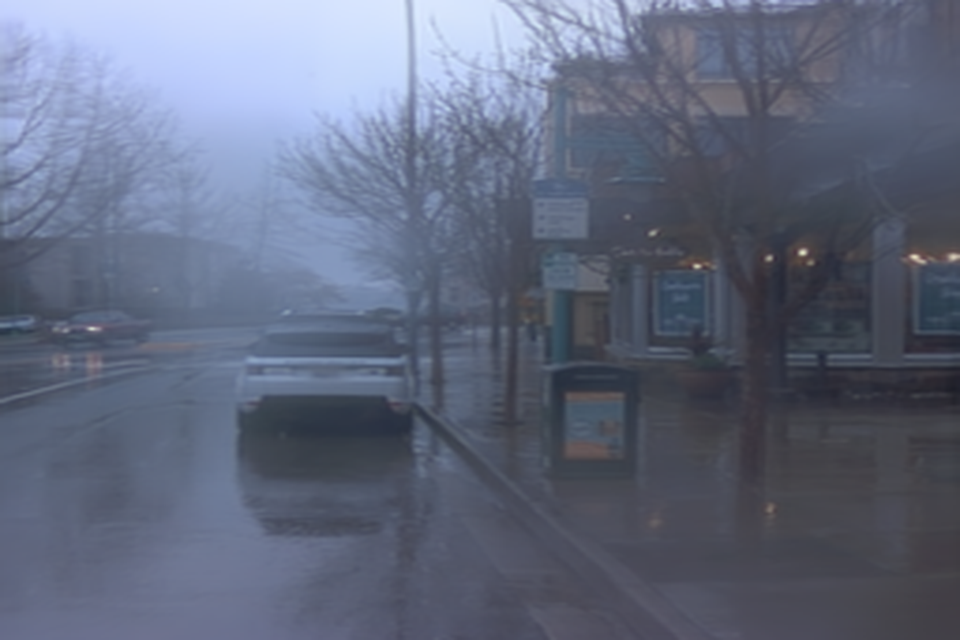} &
    \includegraphics[width=0.16\linewidth]{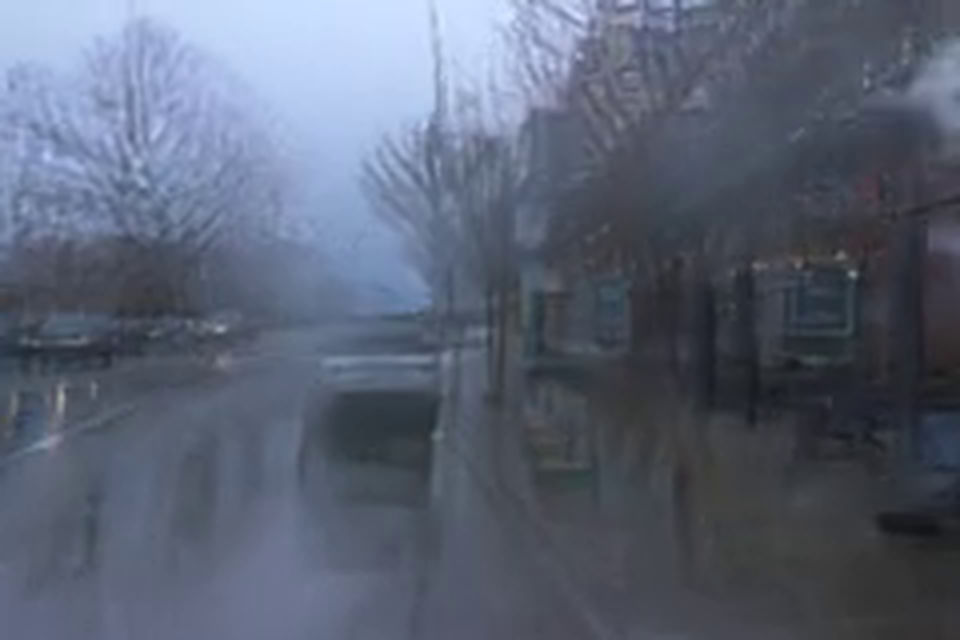} &
    \includegraphics[width=0.16\linewidth]{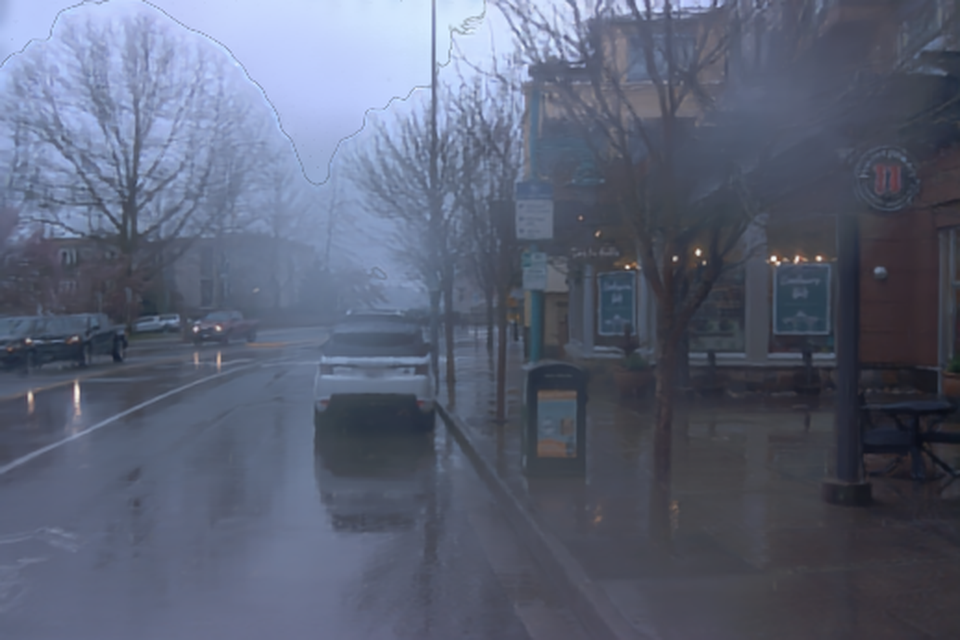} \\ [-0.5mm]
  \end{tabular}

  \caption{\textbf{Qualitative comparison of different methods on Waymo dataset.} (results shown are for the forward-facing camera) }
  \label{fig:comparison_waymo}
\end{figure*}

\subsection{Diffusion-Based Interpolation Refinement} \label{fix}

\textbf{Motivation.} Although the estimated motion field provides a plausible description of object dynamics, interpolation still produces artifacts such as ghosting and disocclusion gaps (Fig .\ref{fig:edit}), due to deviations in the estimated motion and the limited robustness of the 3DGS representation under large rotations and translations. To enhance realism, we further introduce an image-space refinement stage using a diffusion model. This post-render module aims to suppress interpolation artifacts and produce high-quality renderings.

\textbf{Network.} We build our refinement model $f_{\text{diffusion}}$ on a single-step diffusion framework~\cite{sauer2024adversarial}, consisting of a frozen VAE encoder, a UNet denoiser, and a LoRA fine-tuned decoder. As illustrated in Fig.~\ref{fig:main}, given a rendered image $\hat{I}^{t_i}$ and a reference image $I_\text{ref}$ randomly sampled from the input sequence, we first concatenate them frame-wisely and encode the result into the latent space using the VAE encoder. The latent embedding is subsequently processed by the UNet denoiser and decoded to yield the refined image $\tilde{I}^{t_i}$. Formally, the process is formulated as:
\begin{align}
\tilde{I}^{t_i} = f_{\text{diffusion}}(\hat{I}^{t_i}, I_\text{ref}).
\end{align}

\textbf{Loss.} We curated approximately 2,000 clips from 798 Waymo training scenes to train our diffusion-based refinement module. The inputs to the module are interpolated frames containing artifacts, and the outputs are the corresponding refined frames. This dataset provides the supervision necessary for the module to effectively suppress artifacts and produce visually realistic interpolations. The diffusion model is supervised with an $\ell_2$ reconstruction loss between the model output $\tilde{I}^{t_i}$ and the ground-truth image ${I}^{t_i}$, complemented by a perceptual LPIPS loss. To further enhance sharpness and fine details, we also incorporate a style loss based on the Gram matrices of VGG-16 features. The overall objectives is defined as:
\begin{align}
\mathcal{L_\text{diffusion}} = \mathcal{L}_{\text{Recon}} + \mathcal{L}_{\text{LPIPS}} + \lambda_{\text{Gram}} \mathcal{L}_{\text{Gram}}.
\end{align}


\section{Experiment}

\subsection{Experimental protocol}

\paragraph{Datasets} We conduct experiments on the Waymo Open Dataset \cite{sun2020scalability}, which contains real-world autonomous driving logs. For training, we use the full Waymo Open Dataset training split, which comprises 798 scenes, each containing 190–200 frames. \textbf{For evaluation}, we use the Waymo Open Dataset test split, which comprises 202 scenes and covers challenging scenarios such as multi-vehicle interactions, nighttime driving, and rainy weather.
We assessed the generalization of our method across diverse driving conditions with zero-shot tests and training-based experiments on the nuScenes \cite{caesar2020nuscenes} and Argoverse2 \cite{wilson2023argoverse} datasets. For quantitative evaluation on 4D reconstruction, performance is reported using PSNR and SSIM for images, as well as root mean square error (RMSE) for depth estimation. For 3D tracking evaluation, we EPE3D, Acc5, Acc10, and angular error as metrics, following STORM \cite{yang2024storm}.

\subsection{Comparison with SOTA methods}

\begin{figure*}
  \centering
  \makebox[0.01\linewidth][c]{}
    \makebox[0.05\linewidth][c]{\textbf{Input}}
  \makebox[0.2\linewidth][c]{\textbf{T}}
  \makebox[0.2\linewidth][c]{\textbf{T+1}}
  \makebox[0.01\linewidth][c]{}
  \makebox[0.05\linewidth][c]{\textbf{Input}}
  \makebox[0.2\linewidth][c]{\textbf{T}}
  \makebox[0.2\linewidth][c]{\textbf{T+1}}
    \includegraphics[width=0.95\linewidth]{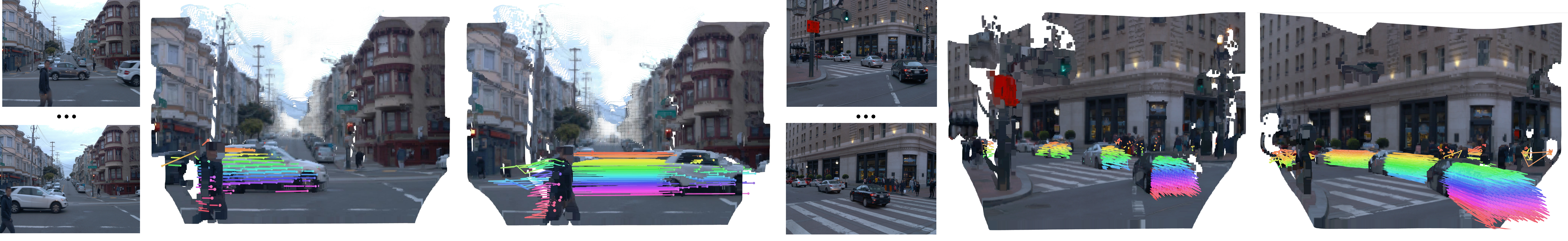}
 \caption{\textbf{3D Tracking Visualization}. Points with the same color correspond across frames.}
  \label{fig:track_qual}
  \vspace{-.5cm}
\end{figure*}

\begin{table*}[t]
  \centering
  \small
  \begin{tabular}{lccc|ccc}
    \toprule
    \multirow{2}{*}{\textbf{Method}} & 
    \multicolumn{3}{c|}{\textbf{nuScenes}} & 
    \multicolumn{3}{c}{\textbf{ Argoverse2}} \\
    \cmidrule(lr){2-4} \cmidrule(lr){5-7}
     & \textbf{PSNR} $\uparrow$ & \textbf{SSIM} $\uparrow$ & \textbf{LPIPS} $\downarrow$ 
     & \textbf{PSNR} $\uparrow$ & \textbf{SSIM} $\uparrow$ & \textbf{LPIPS} $\downarrow$ \\
    \midrule
    \multicolumn{7}{l}{\textbf{Zero-shot}} \\
    MVSplat \cite{chen2024mvsplat} & 17.84 & 0.563 & 0.451 &  18.67 & \underline{0.647} & 0.304\\
    NoPoSplat  \cite{ye2024no} & \underline{19.75} & 0.545 & 0.394 & 22.00 & 0.646 & \underline{0.237} \\
    DepthSplat \cite{xu2025depthsplat} & 19.52 & 0.601 & \underline{0.376} & \underline{22.05} &  0.636 & 0.280 \\
    STORM \cite{yang2024storm} & 17.77 & \underline{0.669} & 0.394 &  20.83 & 0.542  &  0.326  \\
    Ours & \textbf{25.31}  & \textbf{0.794} & \textbf{0.152}  &  \textbf{26.34}  & \textbf{0.812}  & \textbf{0.155}  \\
    \midrule
    \multicolumn{7}{l}{\textbf{Trained}} \\
    STORM \cite{yang2024storm} & 24.54 & 0.784 &  0.267 & 24.97  & 0.791  &  0.240  \\
    Ours & \textbf{26.63}  &  \textbf{0.813} & \textbf{0.122}  &  \textbf{26.96}  &  \textbf{0.831} & \textbf{0.118}  \\
    \bottomrule
  \end{tabular}
  \caption{\textbf{Quantitative Comparison under Trained and Zero-Shot Settings on nuScenes and Argoverse2 datasets.} Our Waymo-trained model shows generalization ability in zero-shot NVS on other datasets, with training on each target dataset further boosting performance.}
\vspace{-.3cm}
  \label{tab:nuscenes_argoverse_comparison}
\end{table*}



\paragraph{Baselines} We compare our approach with a broad set of methods, including optimization-based techniques such as EmerNeRF~\cite{yang2023emernerf}, 3DGS~\cite{kerbl20233d}, PVG~\cite{chen2023periodic}, and DeformableGS~\cite{yang2024deformable}, as well as feedforward methods including LGM~\cite{tang2024lgm}, GS-LRM~\cite{zhang2024gs}, MVSplat~\cite{chen2024mvsplat}, NoPoSplat~\cite{ye2024no}, DepthSplat~\cite{xu2025depthsplat}, and STORM~\cite{yang2024storm}. We further include VGGT++, an extended version of VGGT~\cite{wang2025vggt}, which adds a Gaussian Head for image-space rendering. For tracking, we compare with NSFP~\cite{li2021neural}, NSFP++~\cite{najibi2022motion}, and STORM~\cite{yang2024storm}.

\paragraph{Novel View Synthesis} To evaluate both reconstruction on input frames and novel view synthesis on interpolated frames, we use 20-frame, three-view sequences from 202 scenes as the test set, taking 4 frames as input to predict the intermediate frames and evaluating performance on all frames.  Tab.~\ref{tab:4d recon} reports results on the Waymo dataset, where our method outperforms all baselines. Note that the degraded performance of VGGT++ is due to the limited RGB detail preserved in attention features. Fig.~\ref{fig:comparison_waymo} further provides qualitative comparisons with feedforward approaches. While MVSplat, NoPoSplat, and DepthSplat fail to capture dynamic components due to their limited design, STORM also struggles to accurately model object motion. In contrast, our model successfully distinguishes moving objects from the static background and produces faithful reconstructions. To further assess generalization, we evaluate on nuScenes and Argoverse2 (Tab.~\ref{tab:nuscenes_argoverse_comparison}) under both zero-shot and training-from-scratch settings. In both cases, our model achieves the best performance, demonstrating strong adaptability and robustness across diverse driving scenarios. 

\textit{\textbf{Remark}:} We credit this generalization ability to the pose-free design, which helps reduce domain gaps between datasets.  without relying on explicit camera poses, the model avoids overfitting to dataset-specific trajectory patterns or camera configurations, reducing domain gaps and ensuring robust performance across diverse datasets.
\vspace{-.2cm}

\paragraph{3D Motion Estimation} We evaluate motion estimation on the Waymo Scene Flow dataset. As summarized in Tab.~\ref{tab:storm}, our approach consistently outperforms prior arts across all metrics. These results demonstrate that reliable correspondences can be effectively learned through the rendering loss. Fig.~\ref{fig:track_qual} presents qualitative 3D tracking results at timestamps $T_0$ and $T_1$. We render the 3D motion vectors onto the point clouds, where color-coded tracks highlight consistent correspondences across frames. These results show our method accurately recovers dynamic trajectories for vehicles and pedestrians, validating the motion estimation model.


\subsection{Ablation study and Application}

\begin{table*}[t]
  \centering
  \small
  \begin{tabular}{l c ccc ccc}
    \toprule
    \multirow{2}{*}{\textbf{Method}} & \multirow{2}{*}{\textbf{\#Views Input}} 
    & \multicolumn{3}{c}{\textbf{Reconstruction}} 
    & \multicolumn{3}{c}{\textbf{NVS}} \\
    \cmidrule(lr){3-5} \cmidrule(lr){6-8}
     & & {\small PSNR $\uparrow$} & {\small SSIM $\uparrow$} & {\small D-RMSE $\downarrow$}
       & {\small PSNR $\uparrow$} & {\small SSIM $\uparrow$} & {\small D-RMSE $\downarrow$} \\
    \midrule
    STORM \cite{yang2024storm} & \multirow{2}{*}{4}  & 26.55 & 0.851 & 6.139 & 26.05 & 0.819 & 5.914 \\
    Ours  &   & \textbf{30.54} & \textbf{0.884} & \textbf{3.352} & \textbf{27.41} & \textbf{0.846} & \textbf{3.471} \\
    \midrule
    STORM \cite{yang2024storm} & \multirow{2}{*}{8}  & 25.11 & 0.807 & 6.054 & 25.44 & 0.807 &  5.470 \\
    Ours  &   & \textbf{31.41} &\textbf{ 0.895} & \textbf{3.315} & \textbf{27.74} & \textbf{0.858} & \textbf{3.455} \\
    \midrule
    STORM \cite{yang2024storm} & \multirow{2}{*}{16} & 23.69 & 0.765 & 5.700 & 22.98 & 0.723 & 5.836 \\
    Ours  &   & \textbf{ 30.66} & \textbf{ 0.887} & \textbf{ 3.337} & \textbf{ 28.14} & \textbf{ 0.885} & \textbf{3.480} \\
    \bottomrule
  \end{tabular}%
  \caption{\textbf{Ablation study on the number of input views.} Reconstruction performance shows low sensitivity to the number of input frames, demonstrating robustness with sparse inputs.}
  \label{tab:waymo_comparison}
\end{table*}

\begin{table*}[t]
  \centering
  \scriptsize %
  \setlength{\tabcolsep}{2pt}%

  \begin{minipage}[t]{0.34\textwidth}
    \centering
    \begin{tabularx}{\linewidth}{l *{3}{>{\centering\arraybackslash}X}}
      \toprule
      \textbf{Method}  & \mbox{\textbf{PSNR} $\uparrow$} & \mbox{\textbf{SSIM} $\uparrow$} & \mbox{\textbf{LPIPS} $\downarrow$}  \\
      \midrule
      w/o lifespan & 24.21  & 0.774 & 0.169  \\
      w/o diffusion & \underline{27.32}  & \underline{0.844} &  \textbf{0.108} \\
      Ours  & \textbf{27.41} & \textbf{0.846} &  \underline{0.109} \\
      \bottomrule
    \end{tabularx}
    \captionof{table}{\textbf{Ablation study.} Removing lifespan parameters or the diffusion refinement model decreases performance.}
    \label{tab: 4}
  \end{minipage}\hfill
  \begin{minipage}[t]{0.64\textwidth}
    \centering
    \begin{tabularx}{\linewidth}{l *{4}{>{\centering\arraybackslash}X}}
      \toprule
      \textbf{Method}  & \mbox{\textbf{EPE3D (m)} $\downarrow$} & \mbox{\textbf{Acc$_5$ (\%)} $\uparrow$} & \mbox{\textbf{Acc$_{10}$ (\%)} $\uparrow$} & \mbox{\mbox{\textbf{$\theta$ (rad)} $\downarrow$}} \\
      \midrule
      NSFP      &0.698      & 42.17      &54.26     & 0.919    \\
      NSFP++        &0.711   & 53.10   & 63.02   &0.989    \\
      STORM    & \underline{0.276}    & \underline{81.12}   & \underline{85.61}  & \underline{0.658}  \\
      Ours            &  \textbf{0.183}  &  \textbf{85.42}   &  \textbf{90.42}  &  \textbf{0.328}  \\
      \bottomrule
    \end{tabularx}
    \captionof{table}{\textbf{Quantitative results on 3D motion estimation.} We achieve state-of-the-art performance in 3D tracking on the Waymo Open Dataset.}
    \label{tab:storm}
  \end{minipage}
\end{table*}

\begin{figure*}[!t]
  \centering
  \noindent
  \begin{tabular}{@{}c@{\hspace{0pt}}c@{\hspace{0pt}}c@{\hspace{0pt}}c@{\hspace{0pt}}c@{}}
    \begin{tabular}{@{}c@{}}
      \makebox[0pt][c]{\tiny\textbf{Input Image}} \\[0.5ex]
      \includegraphics[width=0.19\textwidth,height=5cm,keepaspectratio]{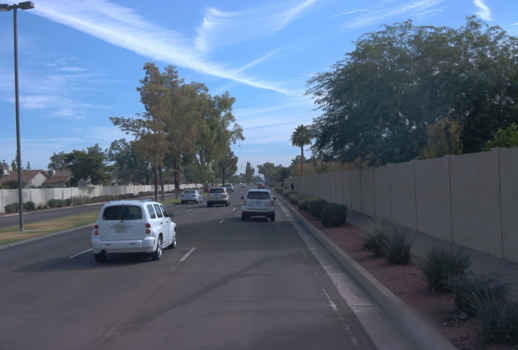}
    \end{tabular}
    &
    \begin{tabular}{@{}c@{}}
      \makebox[0pt][c]{\tiny\textbf{Car Removal (w/o diffusion)}} \\[0.5ex]
      \includegraphics[width=0.19\textwidth,height=5cm,keepaspectratio]{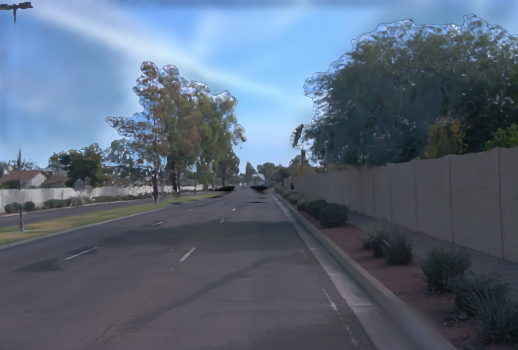}
    \end{tabular}
    &
    \begin{tabular}{@{}c@{}}
      \makebox[0pt][c]{\tiny\textbf{Car Removal(w/ diffusion)}} \\[0.5ex]
      \includegraphics[width=0.19\textwidth,height=5cm,keepaspectratio]{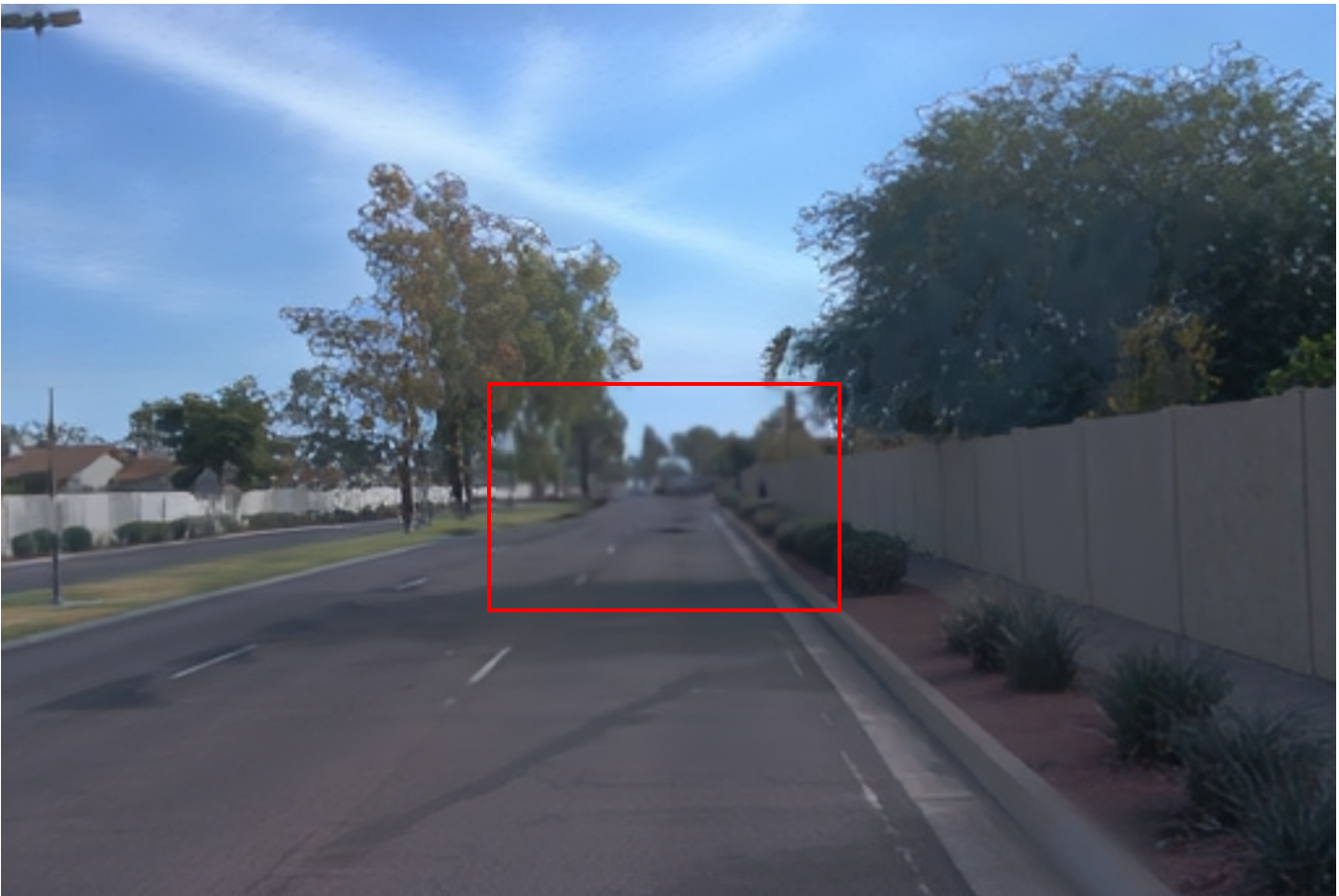}
    \end{tabular}
    &
    \begin{tabular}{@{}c@{}}
      \makebox[0pt][c]{\tiny\textbf{Car Shifting (w/o diffusion)}} \\[0.5ex]
      \includegraphics[width=0.19\textwidth,height=5cm,keepaspectratio]{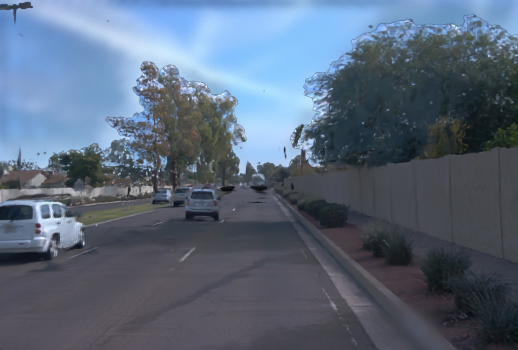}
    \end{tabular}
    &
    \begin{tabular}{@{}c@{}}
      \makebox[0pt][c]{\tiny\textbf{Car Shifting(w/ diffusion)}} \\[0.5ex]
      \includegraphics[width=0.19\textwidth,height=5cm,keepaspectratio]{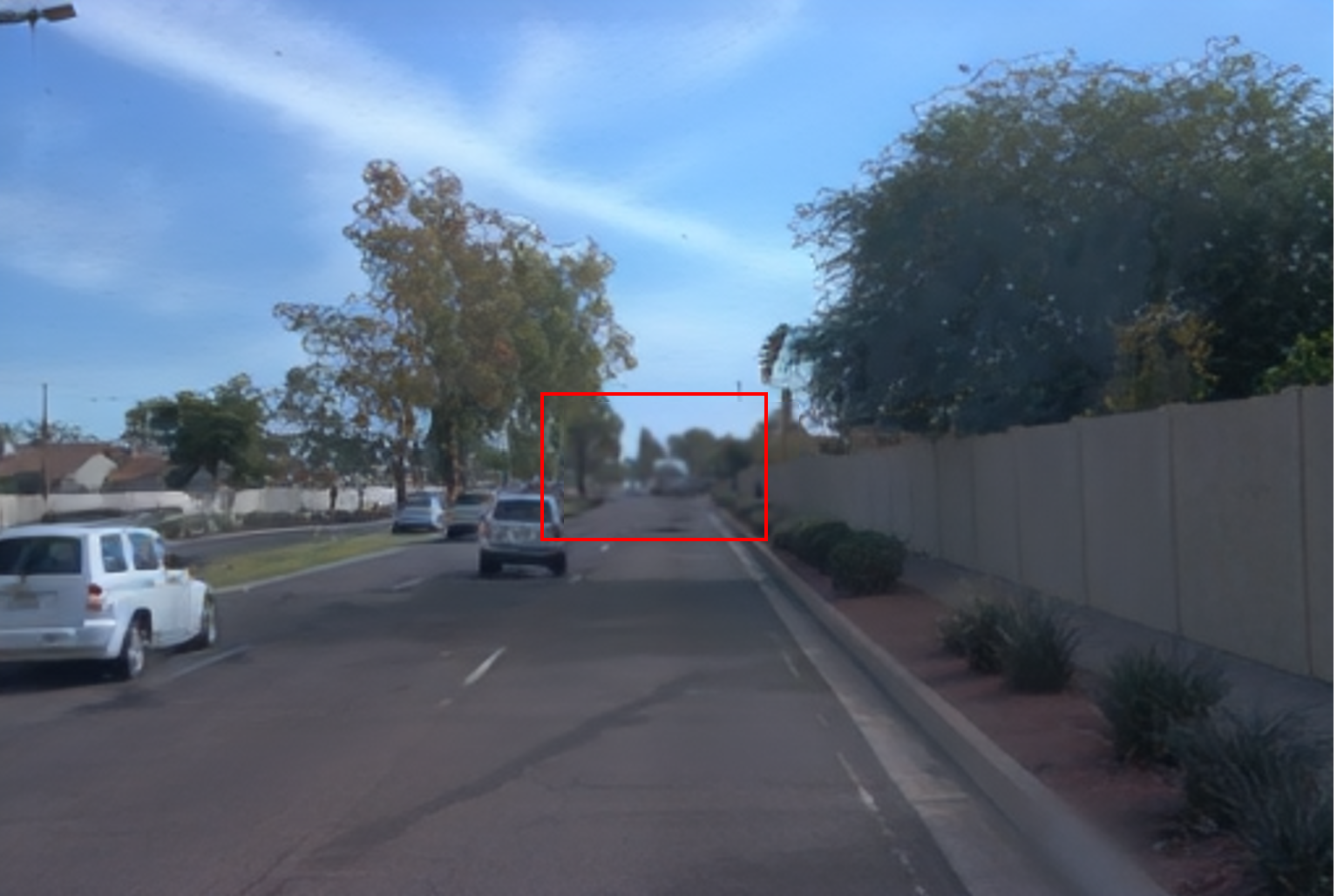}
    \end{tabular}
    \\[1.2\baselineskip]

    \begin{tabular}{@{}c@{}}
      \makebox[0pt][c]{\tiny\textbf{Input Image}} \\[0.5ex]
      \includegraphics[width=0.19\textwidth,height=5cm,keepaspectratio]{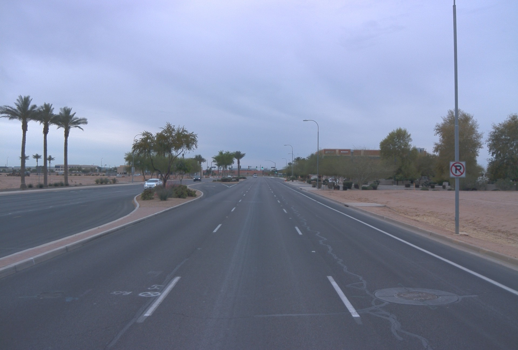}
    \end{tabular}
    &
    \begin{tabular}{@{}c@{}}
      \makebox[0pt][c]{\tiny\textbf{Add Cars (w/o diffusion)}} \\[0.5ex]
      \includegraphics[width=0.19\textwidth,height=5cm,keepaspectratio]{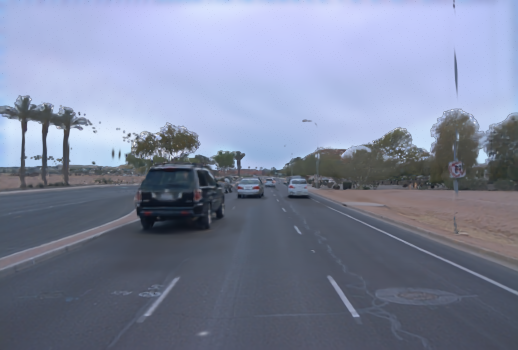}
    \end{tabular}
    &
    \begin{tabular}{@{}c@{}}
      \makebox[0pt][c]{\tiny\textbf{Add Cars(w/ diffusion)}} \\[0.5ex]
      \includegraphics[width=0.19\textwidth,height=5cm,keepaspectratio]{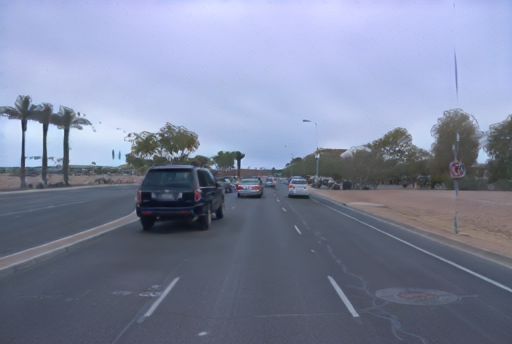}
    \end{tabular}
    &
    \begin{tabular}{@{}c@{}}
      \makebox[0pt][c]{\fontsize{4.5pt}{6pt}\selectfont\textbf{Add Car\&Cyclist (w/o diffusion)}} \\[0.5ex]
      \includegraphics[width=0.19\textwidth,height=5cm,keepaspectratio]{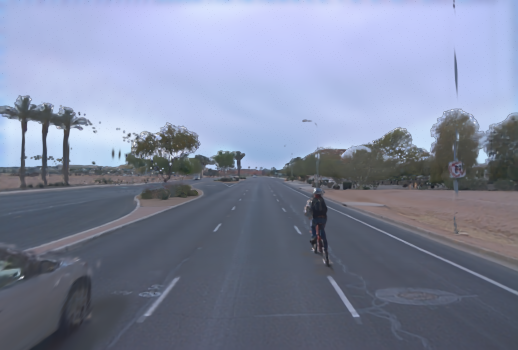}
    \end{tabular}
    &
    \begin{tabular}{@{}c@{}}
      \makebox[0pt][c]{\fontsize{4.5pt}{6pt}\selectfont\textbf{Add Car\&Cyclist(w/ diffusion)}} \\[0.5ex]
      \includegraphics[width=0.19\textwidth,height=5cm,keepaspectratio]{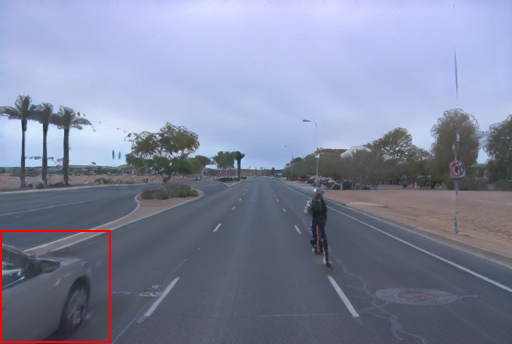}
    \end{tabular}
  \end{tabular}

  \caption{\textbf{Scene editing results}. Cars can be removed or shifted (row 1), and novel vehicles/cyclists inserted from other scenes (row 2). Diffusion refinement fixes artifacts such as holes (red box).}
  \label{fig:edit}
  \vspace{-.6cm}
\end{figure*}

\paragraph{Number of Input Views} One advantage of our method over previous feedforward approaches is its support for a flexible number of input views while maintaining strong performance on long sequences. Tab.~\ref{tab:waymo_comparison} reports both reconstruction results on input frames and NVS results on interpolated frames with 4, 8, and 16 input views. Our approach not only consistently outperforms STORM across all evaluation metrics, but also remains stable as the number of views increases. In contrast, STORM exhibits a sharp drop in PSNR and SSIM as views increase, demonstrating our scalability and robustness under varying input settings.
\vspace{-.3cm}

\begin{figure}[!t]
  \centering
  \begin{picture}(0,0)
  \end{picture}

  \begin{tabular}{c@{\hskip 0.02in} c@{\hskip 0.02in} c@{\hskip 0.02in} c@{\hskip 0.02in} c@{\hskip 0.02in} c@{\hskip 0.02in}}
    \scriptsize\textbf{GT} & \scriptsize\textbf{w/o lifespan} & \scriptsize\textbf{w/ lifespan} & \scriptsize\textbf{GT} & \scriptsize\textbf{w/o diffusion} & \scriptsize\textbf{w/ diffusion} \\
    
    \includegraphics[width=0.16\linewidth]{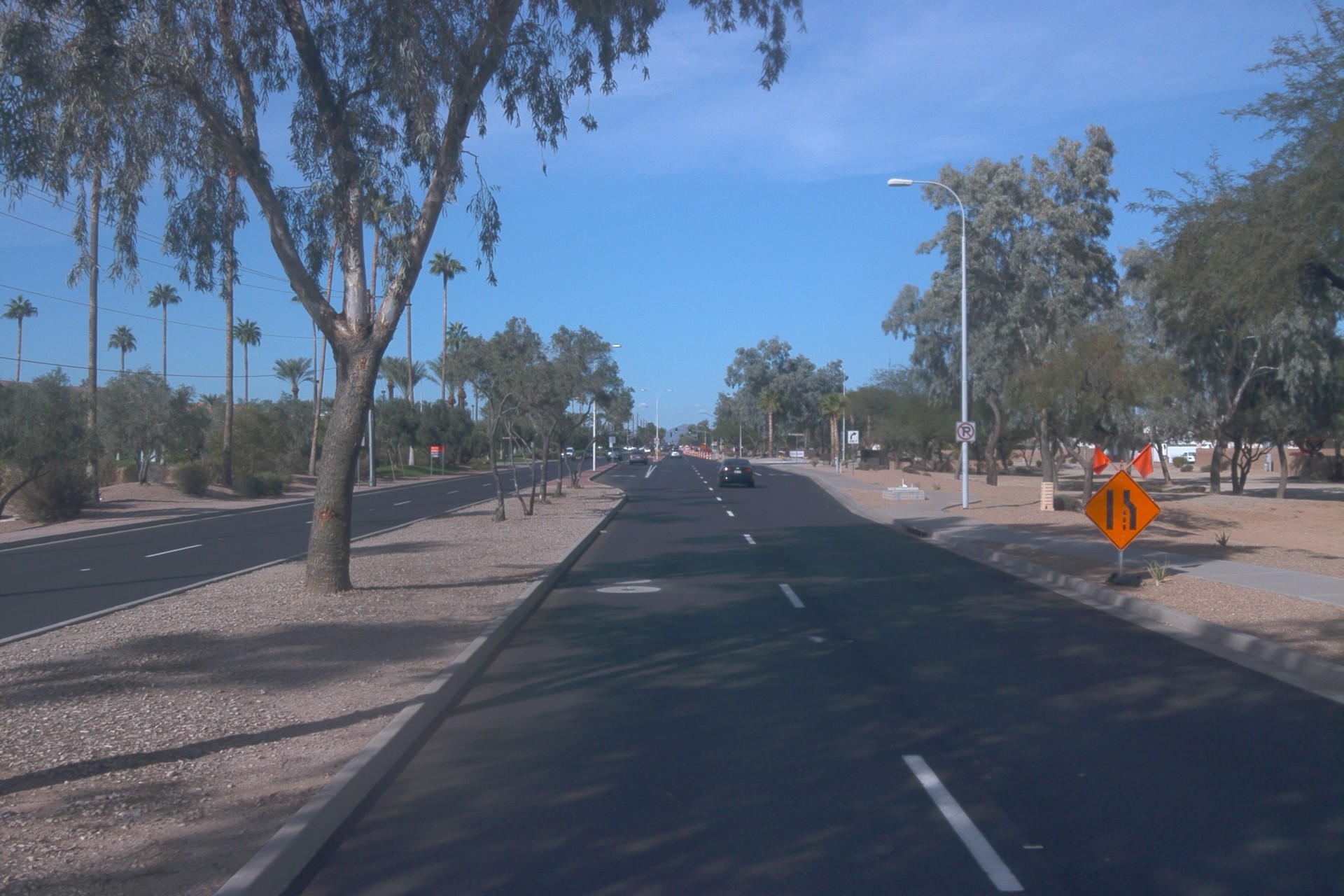} &
    \includegraphics[width=0.16\linewidth]{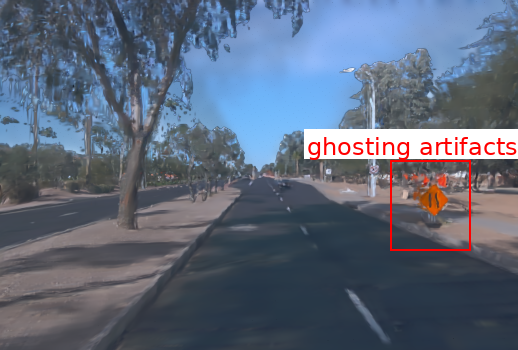} &
    \includegraphics[width=0.16\linewidth]{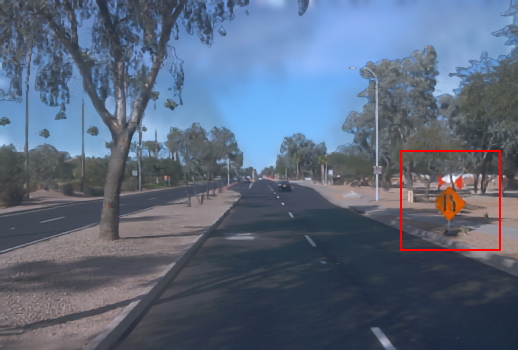} &
    \includegraphics[width=0.16\linewidth]{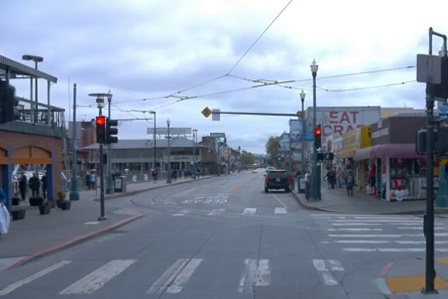} &
    \includegraphics[width=0.16\linewidth]{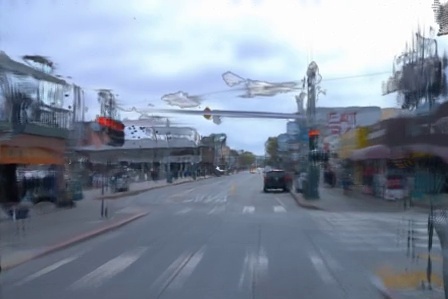} &
    \includegraphics[width=0.16\linewidth]{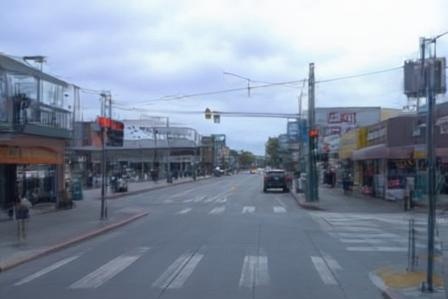} \\ [-1mm]
    
    \includegraphics[width=0.16\linewidth]{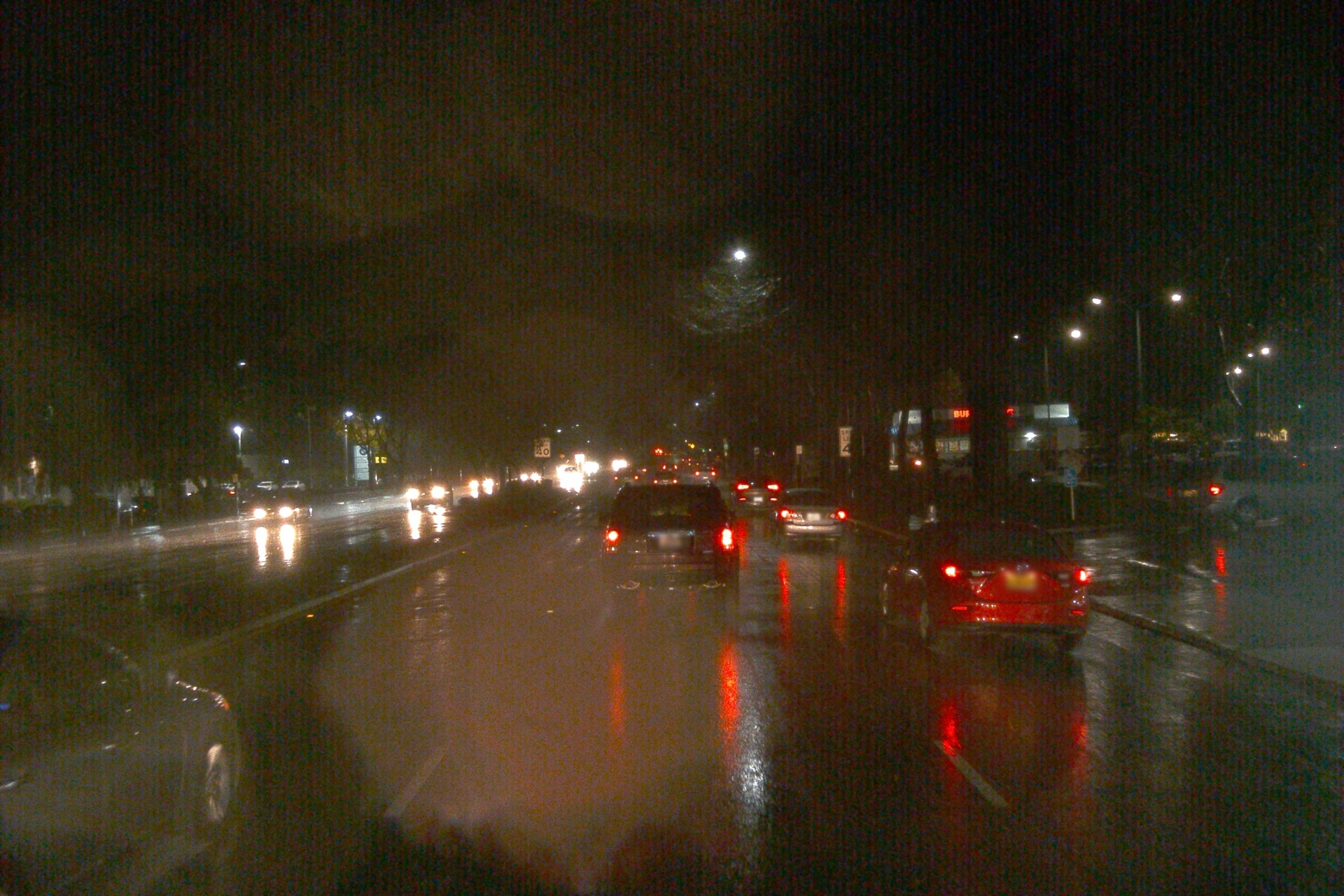} &
    \includegraphics[width=0.16\linewidth]{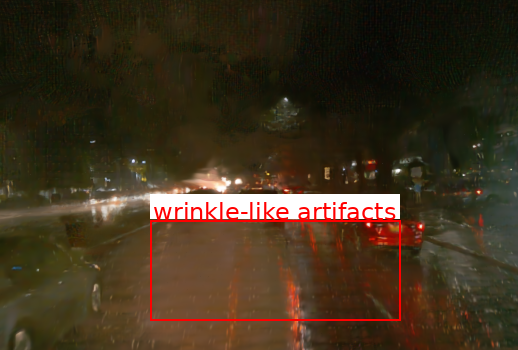} &
    \includegraphics[width=0.16\linewidth]{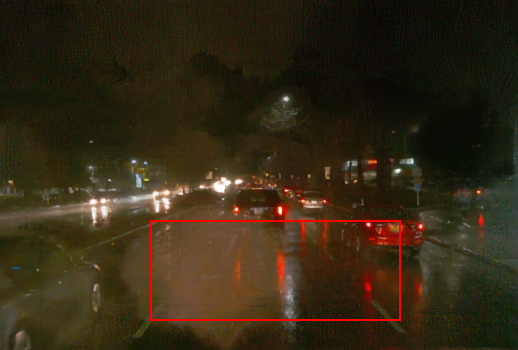} &
    \includegraphics[width=0.16\linewidth]{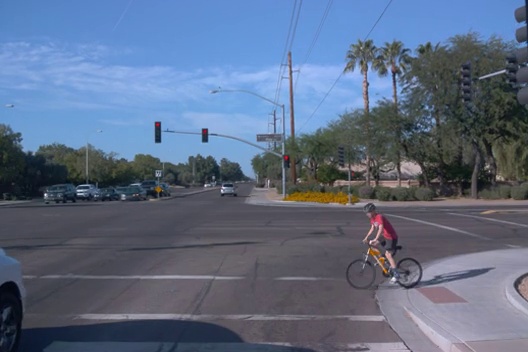} &
    \includegraphics[width=0.16\linewidth]{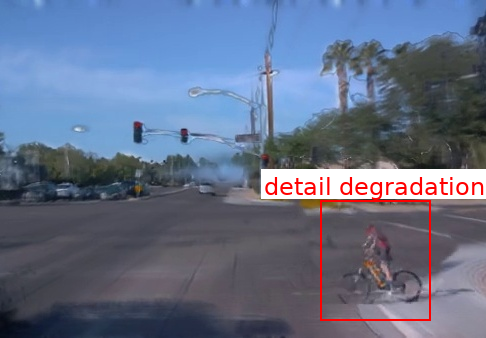} &
    \includegraphics[width=0.16\linewidth]{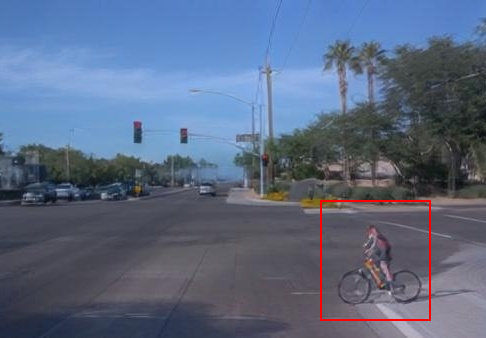} \\ [-0.5mm]
    
    \includegraphics[width=0.16\linewidth]{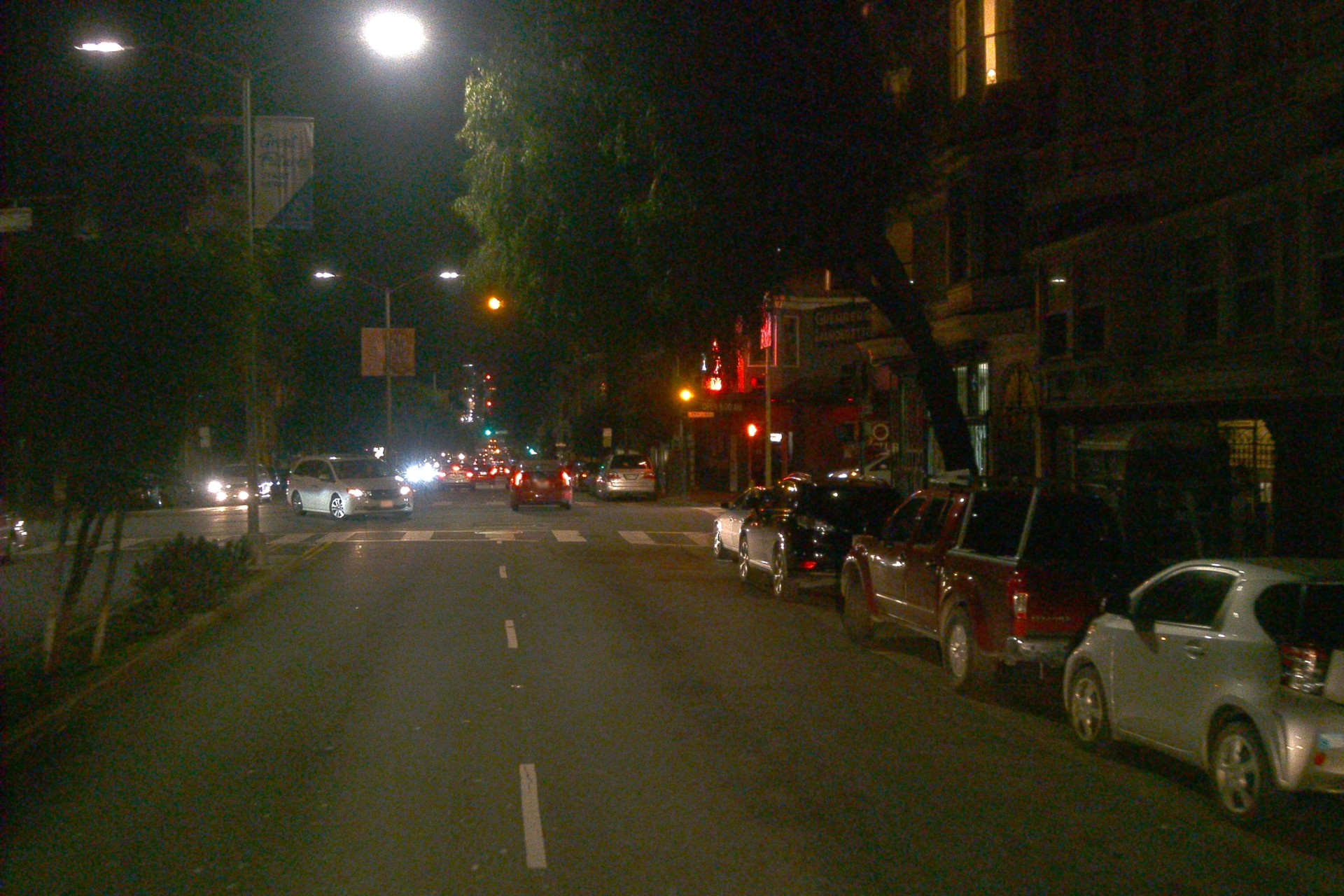} &
    \includegraphics[width=0.16\linewidth]{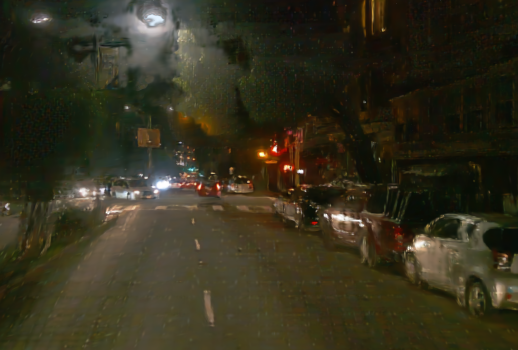} &
    \includegraphics[width=0.16\linewidth]{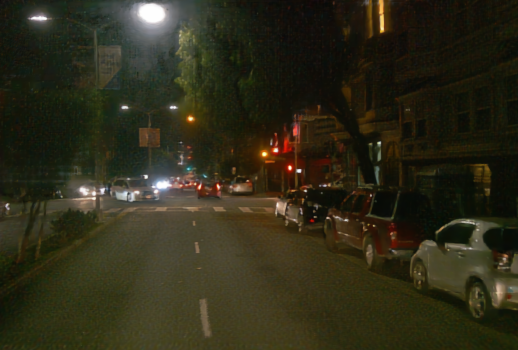} &
    \includegraphics[width=0.16\linewidth]{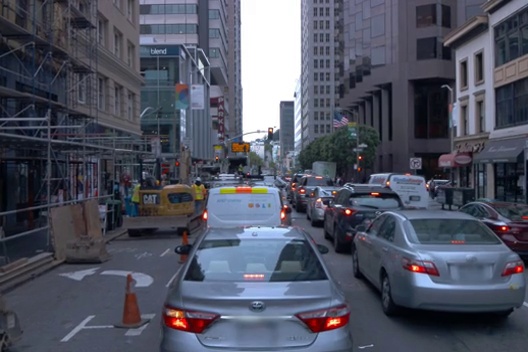} &
    \includegraphics[width=0.16\linewidth]{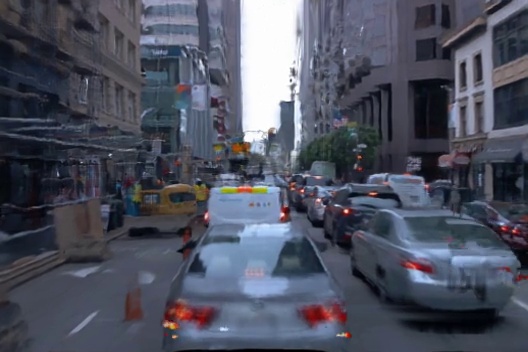} &
    \includegraphics[width=0.16\linewidth]{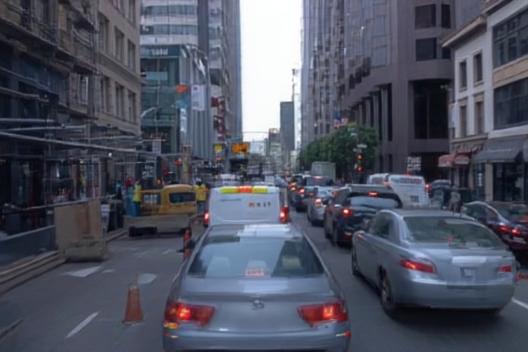} \\ [-0.5mm]
  \end{tabular}
  \caption{\textbf{Ablation study}. Removing the lifespan parameter hinders the capture of changing appearance of static scene, while the diffusion refinement reduces artifacts and improves rendering.}
  \label{fig:difix}
\end{figure}

\paragraph{Lifespan Parameter} We conduct an ablation study to assess the contributions of different model components, summarized in Tab.~\ref{tab: 4} and Fig.~\ref{fig:difix}. Removing the lifespan parameter causes a performance drop, with PSNR decreasing from 27.41 to 24.21. This occurs because lifespan is crucial for capturing the dynamic appearance of static objects, such as lighting changes over time (Fig.~\ref{fig:difix}, row 3). Without it, Gaussians fail to model these subtle variations, resulting in unstable reconstructions and degraded perceptual quality.
\vspace{-.5cm}

\paragraph{Diffusion Refinement} To evaluate the effectiveness of our diffusion refinement module, we conduct an ablation study with and without it, as summarized in Tab.~\ref{tab: 4} and visualized in Fig.~\ref{fig:difix}. Quantitatively, the refinement improves PSNR and SSIM, confirming its role in enhancing rendering quality. Qualitatively, without refinement the reconstructions show noticeable artifacts and loss of fine details, particularly in the sky and dynamic objects. In contrast, incorporating diffusion refinement removes these artifacts and produces more realistic renderings.
We note the diffusion refinement yields only small numeric gains in PSNR/SSIM yet delivers clear qualitative improvements. PSNR/SSIM emphasize per-pixel fidelity, while the diffusion module corrects structural defects and reconstructs missing content, thereby improving downstream usability (e.g., scene editing and robustness to sparse inputs). Consequently, despite modest numeric gains, we retain the diffusion stage.
\vspace{-.2cm}

\paragraph{Scene Editing} We further demonstrate the scene editing capability of our feedforward 4D reconstruction model. Since the method directly generates 3D Gaussian representations and decomposes them into dynamic and static components, we can flexibly manipulate the reconstructed scene by adding, removing, or repositioning dynamic objects. As illustrated in Fig.~\ref{fig:edit}, the first row shows examples where cars are removed or shifted by modifying the dynamic Gaussians, while the second row demonstrates the seamless composition of novel vehicles and a cyclist by integrating the static scene with dynamic Gaussians reconstructed from other scenes. This editing process highlights the advantage of our representation: scene elements can be recombined at the Gaussian level without retraining, enabling efficient modifications of complex driving scenarios. Moreover, the edited images can be further refined using our diffusion-based model, which effectively addresses issues such as holes or artifacts introduced during Gaussian manipulation, as highlighted in the red box.

\section{Conclusions}

We proposed a pose-free, feedforward framework for 4D scene reconstruction from unposed images. Our method unifies camera estimation, 3D Gaussian reconstruction, and 3D motion prediction, while a diffusion-based refinement module improves fidelity under sparse inputs. Experiments on large-scale datasets show SOTA performance with fast speed.  Despite these strengths, some limitations remain. In particular, failure cases can occur when dynamic masks are inaccurate or when tracking fails under heavily occluded motion. Future work will focus on improving dynamic modeling and enhancing tracking robustness to better handle complex dynamic scenes.


{
    \small
    \bibliographystyle{ieeenat_fullname}

}
\clearpage
\setcounter{page}{1}
\maketitlesupplementary


\section{Implementation Details}
\subsection*{Appendix A.1 --- Data Preprocessing}
We construct dynamic masks based on the LiDAR-based 3D bounding-box annotations of the Waymo Open Dataset\cite{sun2020scalability}, which include tracking identifiers. Specifically, we first transform the center of each 3D box from the ego-vehicle coordinate frame to the global coordinate system, and aggregate per-object temporal trajectories using the provided timestamps. Object velocities are then estimated, and category-specific thresholds are applied to determine dynamic instances (pedestrians: $>0.2$\,m/s; vehicles: $>0.5$\,m/s). For objects identified as dynamic, we project their 3D bounding boxes onto the corresponding camera planes to obtain accurate 2D bounding boxes for each frame. These 2D boxes are subsequently used as prompts for an \emph{off-the-shelf} instance segmentation model \cite{ravi2024sam}, combined with temporal information to propagate masks and obtain temporally consistent, object-level instance masks. Static background and sky masks are produced using a semantic segmentation model \cite{xie2021segformer}, whose semantic outputs are further employed in downstream scene understanding tasks.

\subsection*{Appendix A.2 --- Baseline implementations}
\phantomsection\label{sec:appendix.a2}
\paragraph{Baseline selection.}
For all per-scene reconstruction methods shown in Tab.\ref{tab:4d recon}, we selected ten representative baselines. Among them, PVG\cite{chen2023periodic}, DeformableGS\cite{yang2024deformable}, 3DGS\cite{kerbl20233d}, and STORM\cite{yang2024storm} are reconstruction methods implemented on the Waymo Open Dataset\cite{sun2020scalability}; MVSplat\cite{chen2024mvsplat} and DepthSplat\cite{xu2025depthsplat} are feedforward inference networks with strong empirical performance; NoPoSplat\cite{ye2024no}  is a \emph{pose-free} inference framework that does not require camera pose inputs. In addition, we include EmerNeRF\cite{yang2023emernerf}, LGM\cite{tang2024lgm}, and GS-LRM\cite{zhang2024gs} from STORM's reproduced results. For some baselines, reproduced outputs reported by STORM are used directly.

\paragraph{Task setup.}
The task in Tab.\ref{tab:4d recon} and Fig.\ref{fig:comparison_waymo} is short-sequence reconstruction and prediction. 
To align with STORM's experimental setup, we condition on input frames with IDs 0, 5, 10 and 15 and predict frames 0-19 (20 frames in total) -- although our model imposes no restriction on the number or indices of input and output frames. Inference is performed from three camera viewpoints -- the forward-facing, front-left and front-right cameras. 
This represents a basic scene reconstruction and novel view synthesis (NVS) task. Across the ten experiments, methods that require iterative fitting are uniformly trained/fitted for 5,000 iterations; feedforward reconstruction methods are evaluated without this iterative training constraint. All evaluations are performed on the Scene Flow validation split of the Waymo Open Dataset\cite{sun2020scalability}, which contains 202 scenes.

\paragraph{Sky mask and depth metrics.}
Some methods do not explicitly reconstruct the sky; therefore depth-related metrics are computed only over non-sky regions. Sky masks are obtained from Waymo's LiDAR data and filtered to ensure high confidence. For depth evaluation, methods without camera pose input, including NoPoSplat and ours, predict only relative depth, whose scale and offset may differ from ground truth. To ensure fair comparison, we perform a linear alignment of the predictions before computing the error, and then report the aligned depth RMSE (D-RMSE) within the valid mask regions. 

\paragraph{Computation.}
During the training phase, the model was trained on the Waymo Open Dataset\cite{sun2020scalability} using an eight-card H200 GPU configuration. The training process was completed in approximately 24 hours, with convergence achieved at around 5,000 iterations, as indicated by the stabilization of loss values and performance metrics on the validation set.
In the experimental phase, to ensure direct comparability with STORM\cite{yang2024storm} and its reproduced baseline methods, all evaluations in this study were exclusively conducted on NVIDIA A100 GPUs. This hardware alignment guarantees a consistent and fair comparison of computational performance and results across different models.


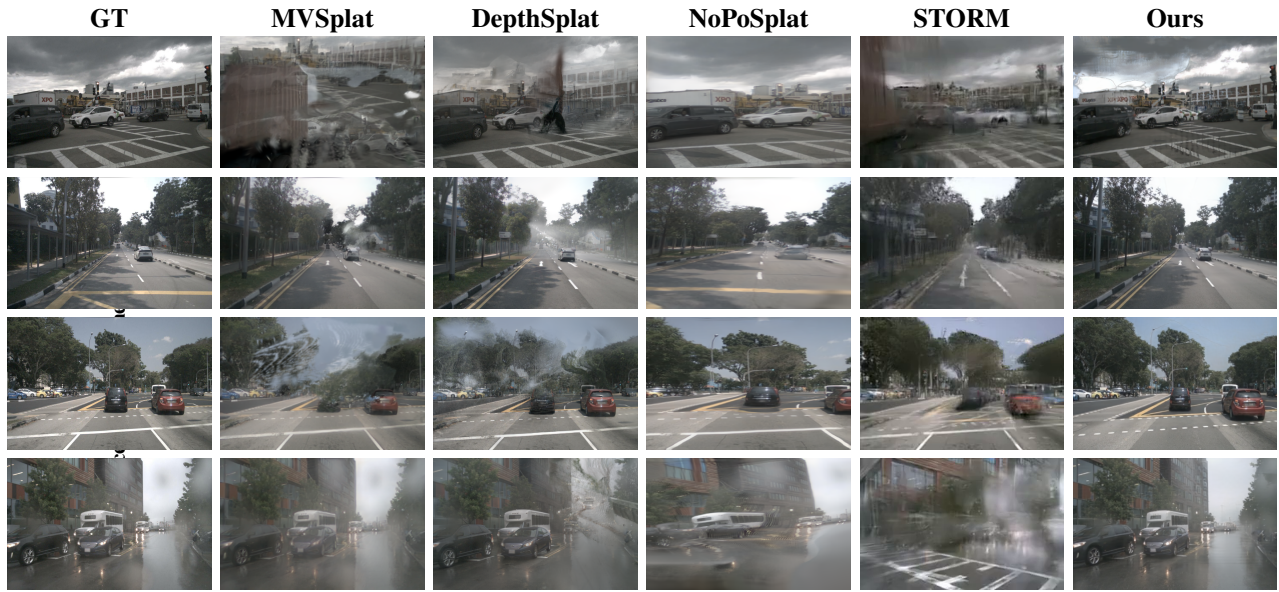
\begin{figure*}[!t]
  \centering
  \begin{picture}(0,0)
    \put(-200, -44){\rotatebox{90}{\footnotesize\textbf{Scene1}}}
    \put(-200, -91){\rotatebox{90}{\footnotesize\textbf{Scene2}}}
    \put(-200, -134){\rotatebox{90}{\footnotesize\textbf{Scene3}}}
    \put(-200, -179){\rotatebox{90}{\footnotesize\textbf{Scene4}}}
  \end{picture}

  \begin{tabular}{c@{\hskip 0.05in} c@{\hskip 0.05in} c@{\hskip 0.05in} c@{\hskip 0.05in} c@{\hskip 0.05in} c@{\hskip 0.05in}}
    \textbf{GT} & \textbf{MVSplat} & \textbf{DepthSplat} & \textbf{NoPoSplat} & \textbf{STORM} & \textbf{Ours} \\
    
    \includegraphics[width=0.155\linewidth,height=0.1\linewidth]{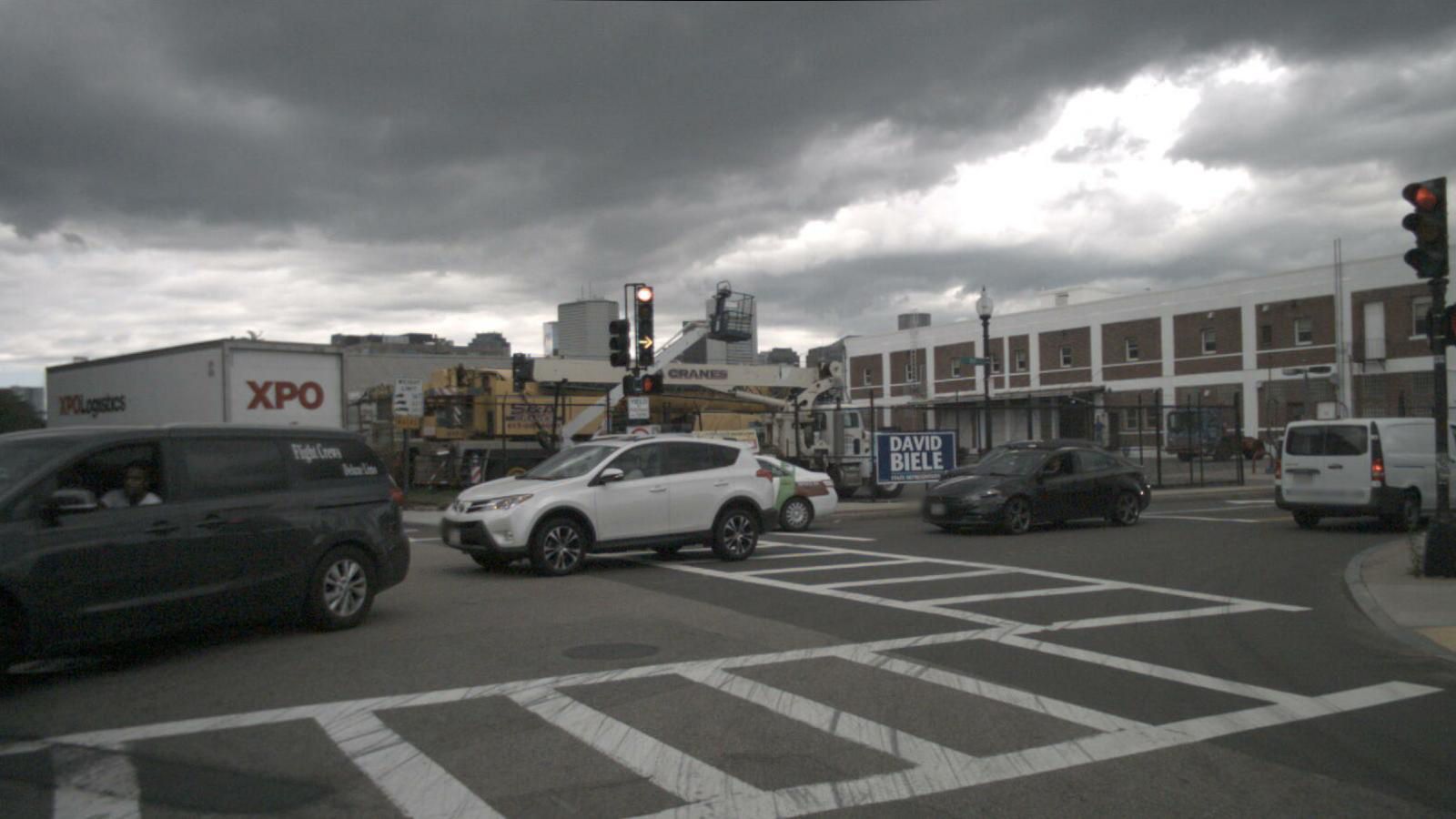} &
    \includegraphics[width=0.155\linewidth,height=0.1\linewidth]{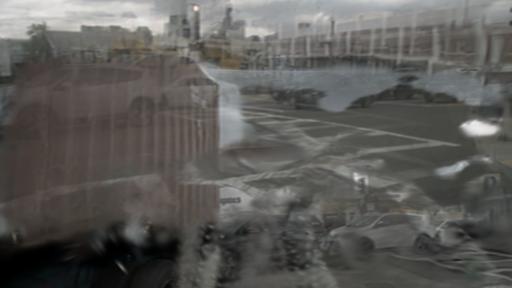} &
    \includegraphics[width=0.155\linewidth,height=0.1\linewidth]{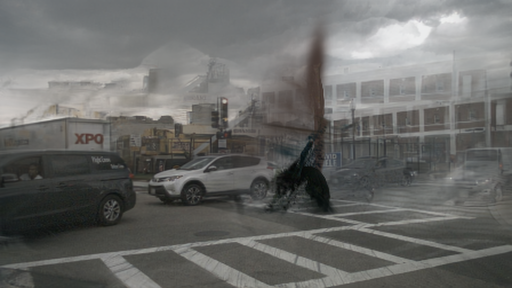} &
    \includegraphics[width=0.155\linewidth,height=0.1\linewidth]{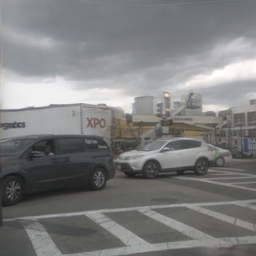} &
    \includegraphics[width=0.155\linewidth,height=0.1\linewidth]{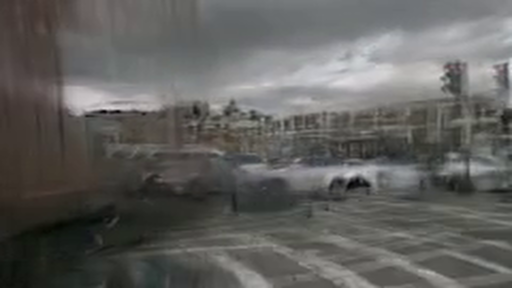} &
    \includegraphics[width=0.155\linewidth,height=0.1\linewidth]{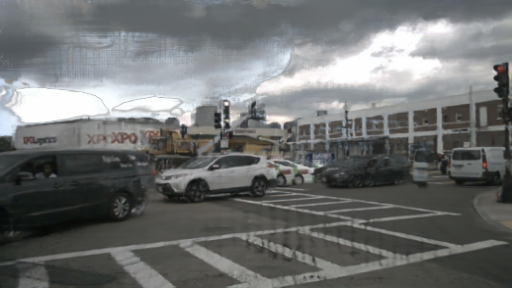} \\
    
    \includegraphics[width=0.155\linewidth,height=0.1\linewidth]{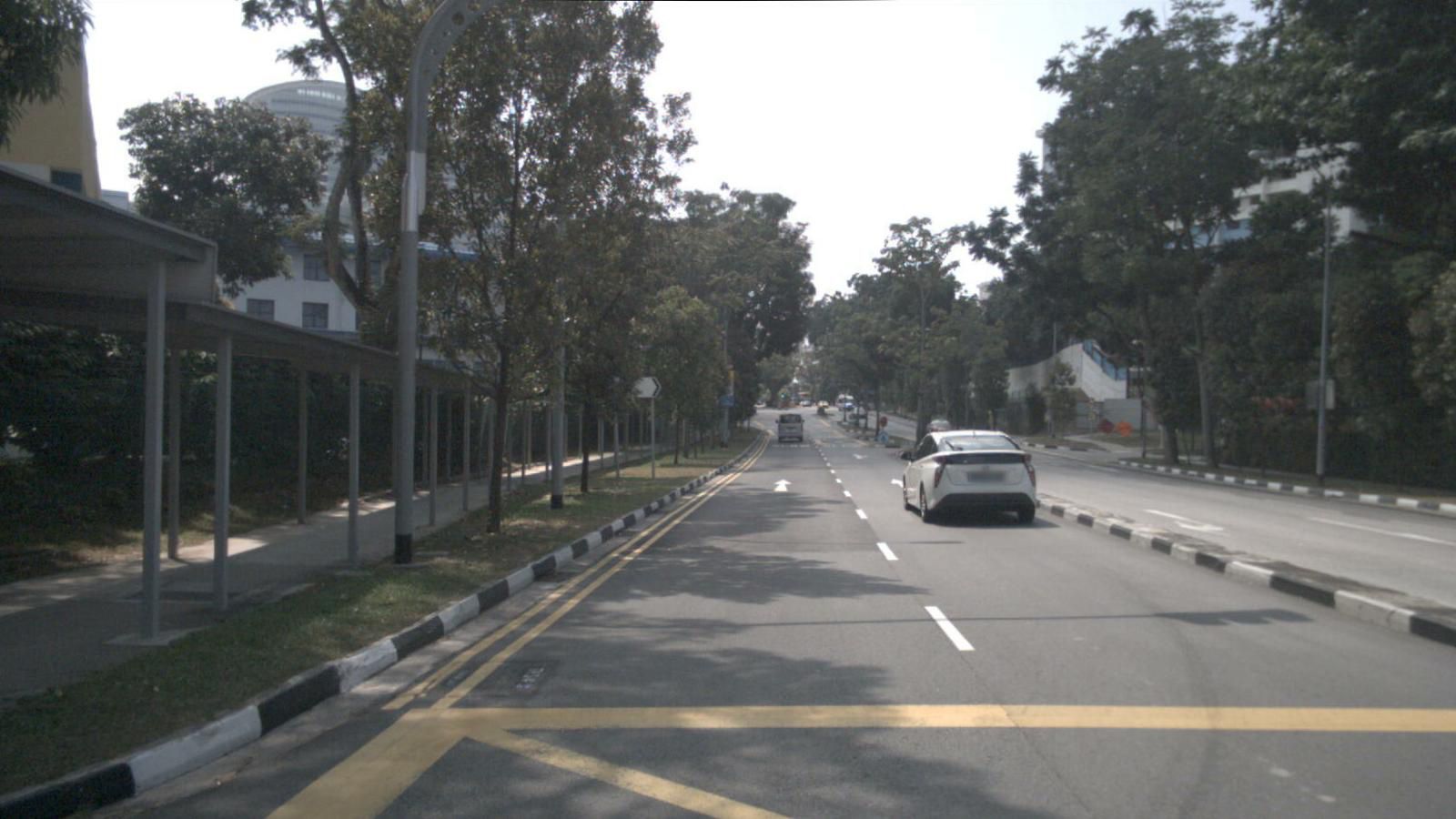} &
    \includegraphics[width=0.155\linewidth,height=0.1\linewidth]{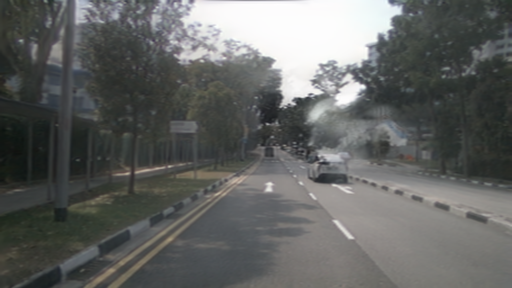} &
    \includegraphics[width=0.155\linewidth,height=0.1\linewidth]{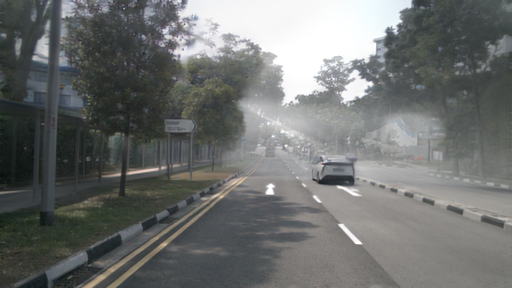} &
    \includegraphics[width=0.155\linewidth,height=0.1\linewidth]{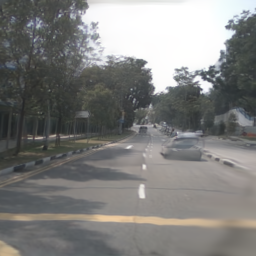} &
    \includegraphics[width=0.155\linewidth,height=0.1\linewidth]{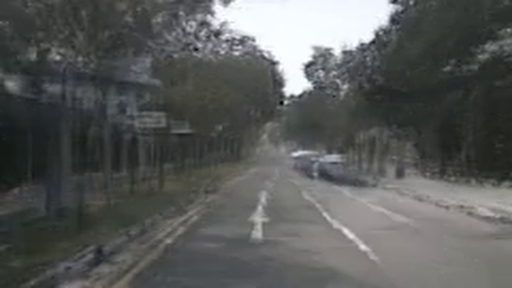} &
    \includegraphics[width=0.155\linewidth,height=0.1\linewidth]{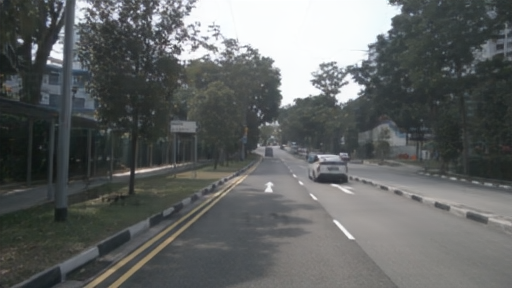} \\
    
    \includegraphics[width=0.155\linewidth,height=0.1\linewidth]{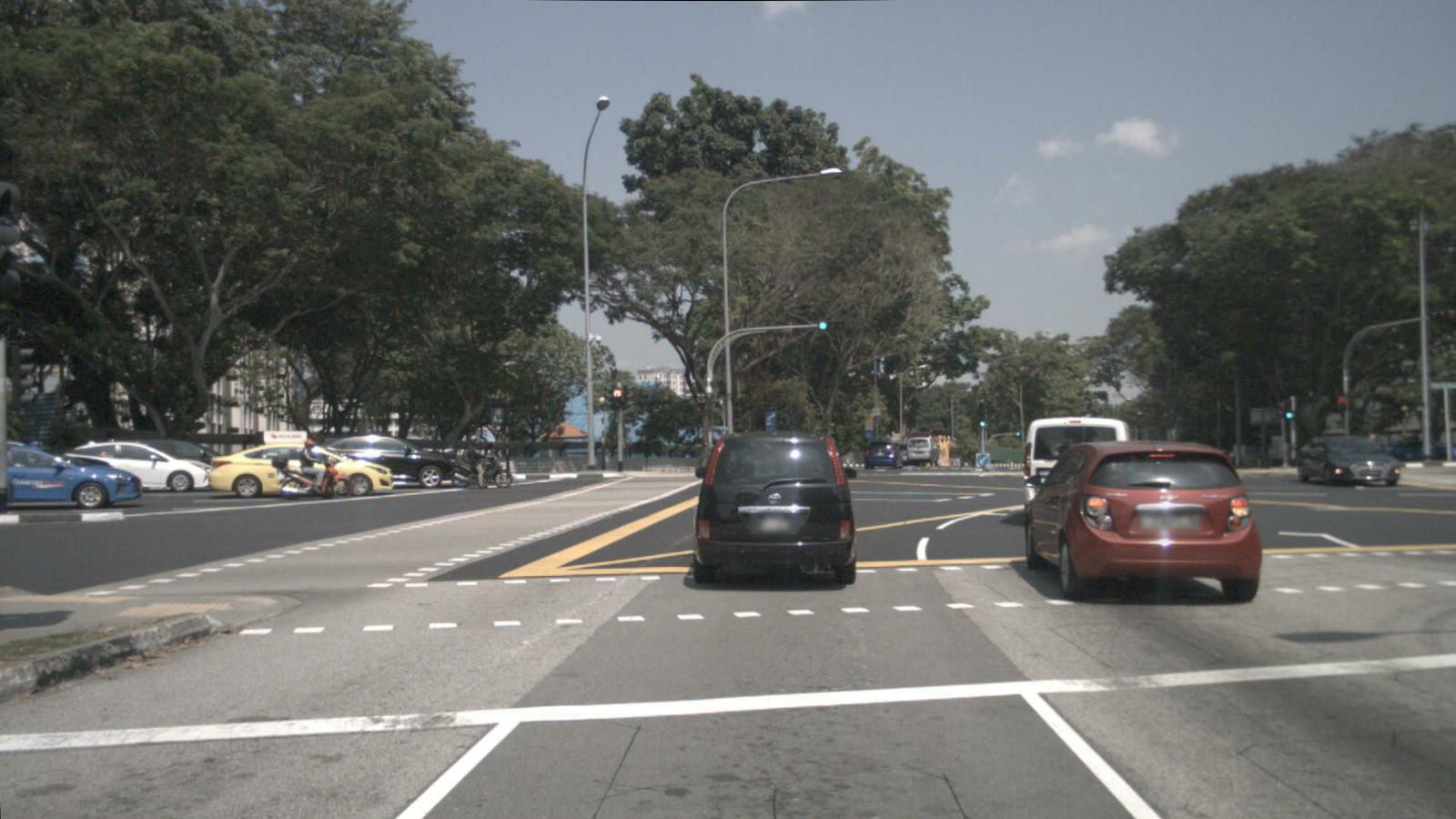} &
    \includegraphics[width=0.155\linewidth,height=0.1\linewidth]{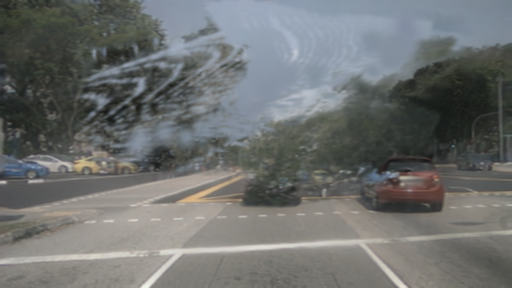} &
    \includegraphics[width=0.155\linewidth,height=0.1\linewidth]{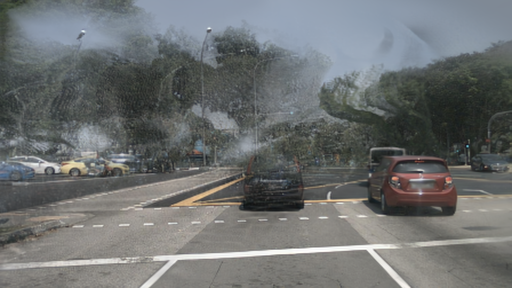} &
    \includegraphics[width=0.155\linewidth,height=0.1\linewidth]{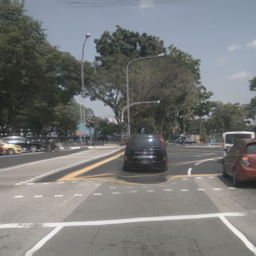} &
    \includegraphics[width=0.155\linewidth,height=0.1\linewidth]{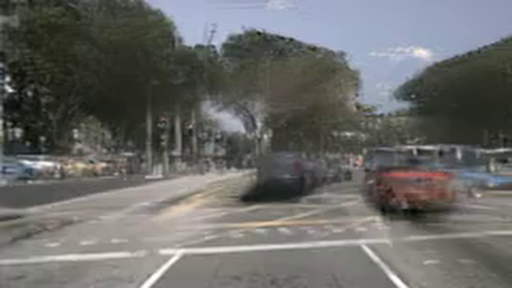} &
    \includegraphics[width=0.155\linewidth,height=0.1\linewidth]{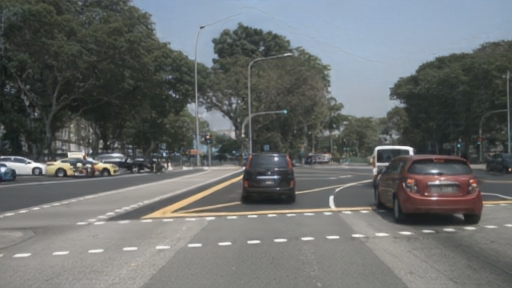} \\
    
    \includegraphics[width=0.155\linewidth,height=0.1\linewidth]{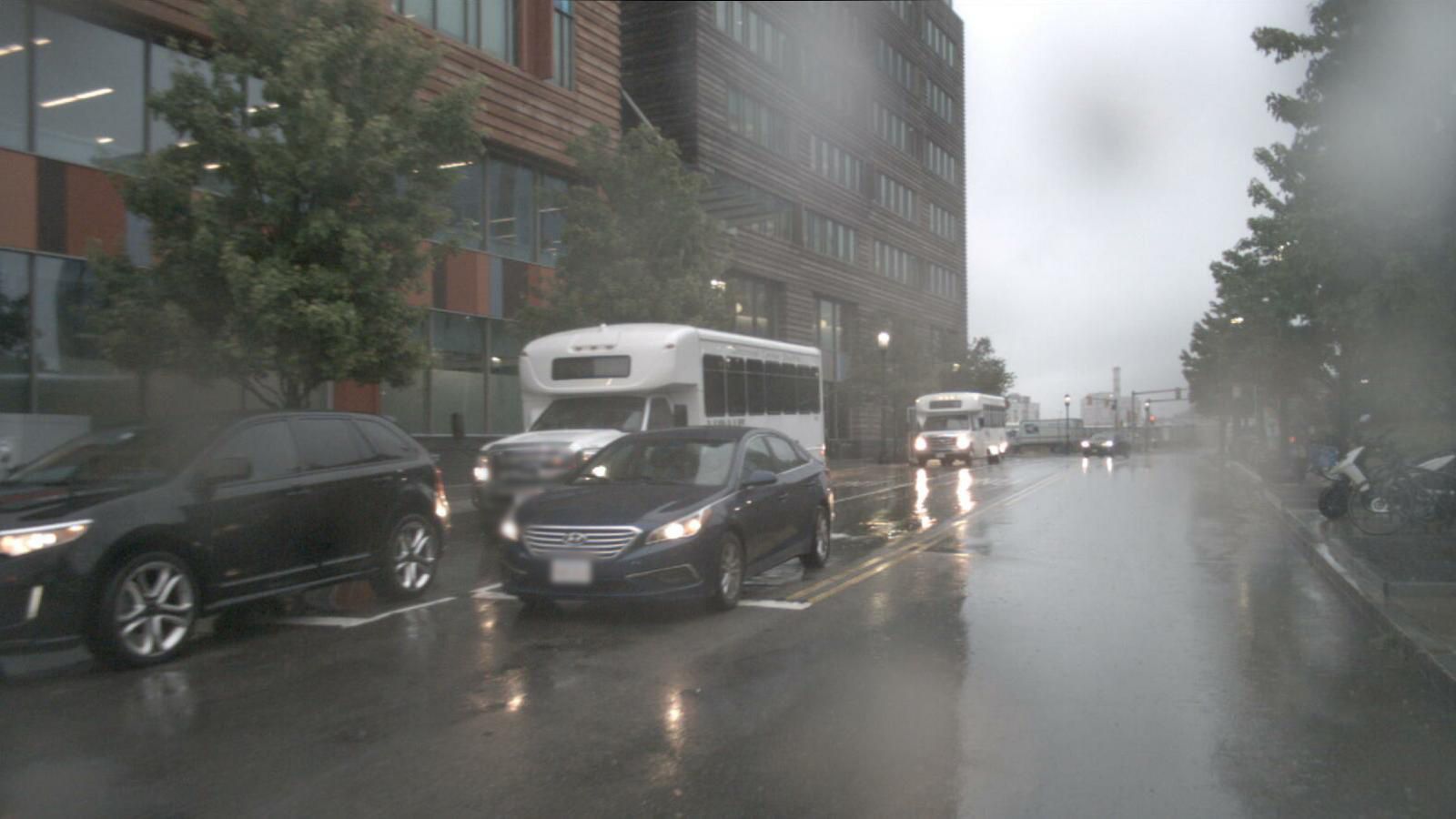} &
    \includegraphics[width=0.155\linewidth,height=0.1\linewidth]{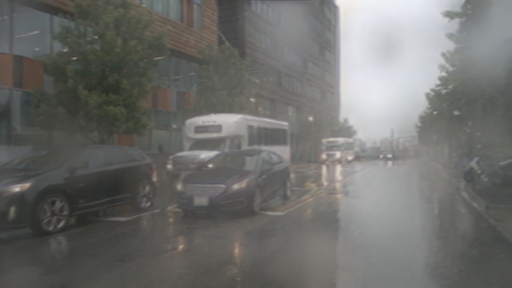} &
    \includegraphics[width=0.155\linewidth,height=0.1\linewidth]{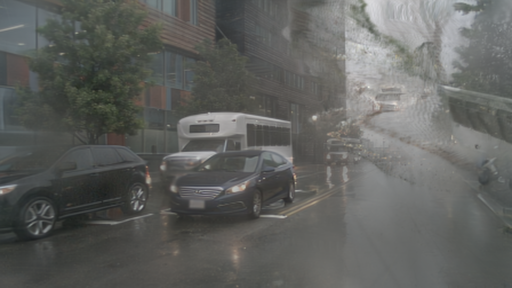} &
    \includegraphics[width=0.155\linewidth,height=0.1\linewidth]{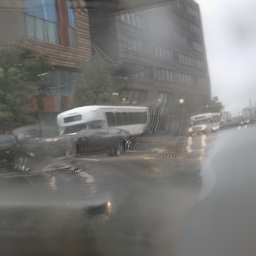} &
    \includegraphics[width=0.155\linewidth,height=0.1\linewidth]{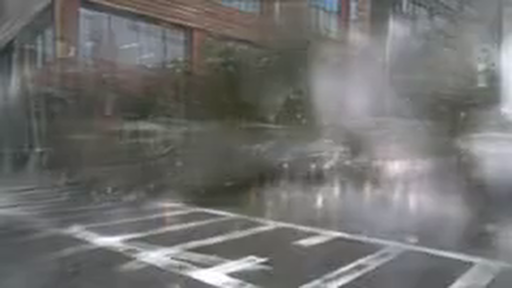} &
    \includegraphics[width=0.155\linewidth,height=0.1\linewidth]{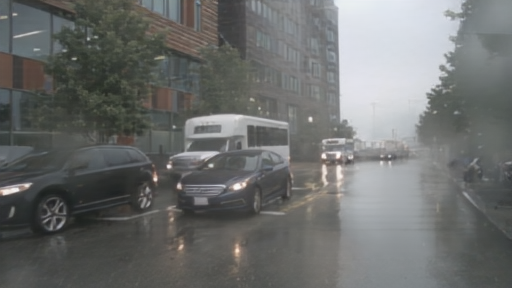} \\
  \end{tabular}
  \caption{Zero-shot experiment on nuScenes and Argoverse2 datasets.}
  \label{fig:zeroshot}
\end{figure*}

\section{Additional Experimental Results}
\subsection*{Appendix B.1 --- Comparison on addtional datasets}
We evaluate the generalization of our model on the public nuScenes \cite{caesar2020nuscenes} and Argoverse2\cite{wilson2023argoverse} datasets. To provide a systematic comparison of cross-domain and in-domain performance, we design two complementary experiments: (1) \textbf{zero-shot evaluation}, where the model is trained on the Waymo Open Dataset\cite{sun2020scalability} and tested directly on nuScenes/Argoverse2 to assess cross-domain generalization;  (2) \textbf{target-domain training evaluation}, where the model is trained and evaluated independently on nuScenes and Argoverse2 to measure upper-bound performance within each target domain.

Sampling and split details are as follows. For nuScenes (v1.0), the dataset contains approximately 1,000 driving scenes, each lasting roughly 20~s, with camera sampling at 12~Hz. We randomly sample 600 scenes for this study, using the first 500 scenes for training and the remaining 100 scenes for testing. For the Argoverse2 Sensor Dataset, which comprises roughly 1,000 annotated driving sequences and provides multi-modal sensor observations, the camera trigger frequency is approximately 20~Hz; we likewise select 600 sequences, with 500 used for training and 100 for evaluation.

To ensure comparability across datasets, all experiments adopt the same preprocessing and training configuration used for Waymo: camera images are uniformly downsampled/resampled to \(518\times518\), and data augmentation, optimizer settings, and training schedules are kept consistent. Models typically converge after roughly \(1{,}000\) epochs. Selected zero-shot inference examples on nuScenes and Argoverse2 are shown in Fig.~\ref{fig:zeroshot}.

    
    
    




\subsection*{Appendix B.2 --- more qualitative results}

\begin{figure}[!t]
  \centering
  {\scriptsize
  \makebox[0.23\linewidth][c]{\textbf{GT}}
  \makebox[0.23\linewidth][c]{\textbf{Rendering}}
  \hspace{1.1em}\makebox[0.23\linewidth][c]{\textbf{Dynamic Rendering}}
  \hspace{0.5em}\makebox[0.23\linewidth][c]{\textbf{Dynamic map}}
  }
  
  \includegraphics[width=0.95\linewidth]{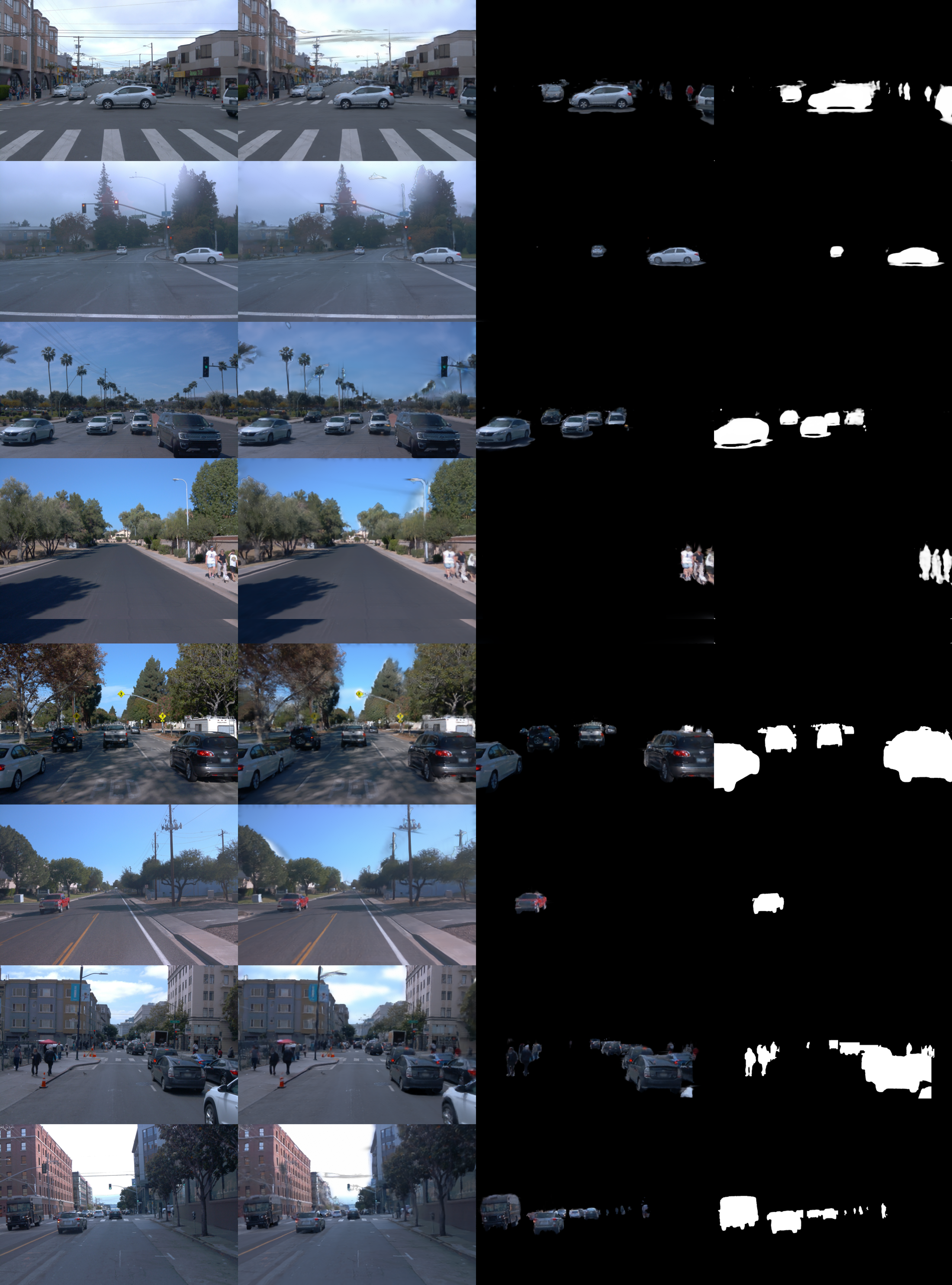}
  
  \caption{More Qualitative Results.}
  \label{fig:more_qual}
\end{figure}

Fig.~\ref{fig:more_qual} presents additional qualitative results of our method, showing full-image renderings, dynamic object renderings, and the predicted dynamic masks. Our approach effectively separates dynamic elements, such as vehicles and pedestrians, from the static background across diverse urban driving scenarios. The dynamic renderings align closely with ground-truth object locations, and the dynamic maps provide accurate object masks, demonstrating the effectiveness of our dynamic scene modeling and motion decomposition.

\end{document}